\documentclass[journal,12pt,onecolumn,draftclsnofoot]{IEEEtran}
\IEEEoverridecommandlockouts

\usepackage{cite}
\usepackage{amsmath,amssymb,amsfonts}
\usepackage{algorithm}
\usepackage{algorithmic}
\usepackage{graphicx,psfrag}
\usepackage{graphicx}
\usepackage{textcomp}
\usepackage{xcolor}
\usepackage{amsthm}
\usepackage[justification=centering]{caption}
\usepackage{overpic}
\usepackage{multirow}
\usepackage{fancyhdr}
\usepackage{lmodern}
\usepackage{subfig}

\title{A Machine Learning Framework for Weighted Least Squares GNSS Positioning based on Activation Functions}
\author{Pin-Hsun~Lee \footnote{Pin-Hsun~Lee is with the Department
of Electrical and Computer Engineering, McGill University, Montreal,
QC, Canada. E-mail:pin-hsun.lee@mail.mcgill.ca.}
 and
 Harry Leib \footnote{Harry Leib is with the Department
of Electrical and Computer Engineering, McGill University, Montreal,
QC, Canada. E-mail:harry.leib@mail.mcgill.ca.}	}

\begin{document}
	\maketitle	

\vspace*{-20mm}

\begin{abstract}
Global Navigation Satellite Systems (GNSS) are widely used to provide position, velocity, and timing (PVT) information for various applications, including transportation, location-based communication services, and intelligent agriculture. In urban canyons, high-rise buildings and narrow streets can cause signal obstruction, non-line-of-sight (NLOS) reception, and multipath effects that introduce errors in GNSS pseudorange measurements. Although multi-constellations GNSS effectively increase the number of available satellites, the inclusion of degraded signals can lead to severe positioning errors. This study proposes a machine learning framework for the weighted least squares (WLS) algorithm incorporating activation functions to enhance positioning accuracy. Several signal quality indicators are employed as training features for ensemble learning algorithms to identify poor quality signals by providing quality scores. Then, activation functions are employed to transform the machine learning predicted scores to appropriate weights for WLS positioning. To evaluate the performance of our approach, experiments are conducted using real-world datasets from Hong Kong and Tokyo urban areas. Comparative analysis of activation functions reveals that sigmoid functions consistently yield the greatest improvements with different machine learning algorithms and GNSS constellation configurations. The proposed algorithm demonstrates substantial reductions in positioning errors for both single- and multi-constellation scenarios. Furthermore, our results indicate that the proposed algorithm exhibits strong geographical transferability. The proposed algorithm maintains comparable level of performance when trained on data from other regions with similar levels of urbanization.

\end{abstract}

\textit{Index Terms} - Global navigation satellite systems (GNSS), Machine learning, Activation functions, Multi-constellations GNSS, Urban canyons, NLOS signal propagation, Multipath signal propagation.

\section{Introduction}
The growing reliance of existing and emerging technologies on position information  prompted an increased interest in  Global Navigation Satellite Systems (GNSS) applications \cite{hegarty2009evolution}\cite{fernandez2011satellite}\cite{egea2022gnss}.  From smartphone navigation \cite{zangenehnejad2021gnss}\cite{siemuri2021improving} to intelligent transportation systems \cite{jin2024survey}\cite{he2023research}, GNSS provides services that are essential to our modern society. With the rapid advancement and commercialization of autonomous vehicles, the importance of GNSS positioning accuracy has reached unprecedented levels \cite{lu2021real}. However, in urban environments GNSS signals are susceptible to obstruction, reflection, and refraction from high-rise buildings which lead to degraded positioning accuracy  \cite{isik_machine_2021}. Signal obstruction results in insufficient number of received signals to generate a positioning solution. Non-line-of-sight (NLOS) signals reflected from building surfaces travel longer paths, which introduce delays in signal arrival time \cite{hsu_analysis_2017}. These propagation delays cause pseudorange measurement errors that can exceed one kilometer \cite{groves_height_2013}. Multipath interference, where the same signal is received via multiple propagation paths, deteriorates the correlation functions and leads to pseudorange measurement errors. \cite{teunissen_springer_2017}. Furthermore, the satellite geometry is a critical factor affecting positioning accuracy and can be quantified by the dilution of precision (DOP) \cite{dussault_influence_2001} metrics. A high DOP indicates poor satellite geometry that leads to higher positioning error \cite{dussault_influence_2001}.

To increase GNSS positioning accuracy, one of the most popular research directions is GNSS NLOS/multipath detection and exclusion \cite{ozeki_gnss_2022,kirmaz_nlos_2021,suzuki_nlos_2021}. The advancement of artificial intelligence (AI) has facilitated the application of machine learning techniques for NLOS/multipath signal classification. AI models can find hidden patterns in complex and high-dimensional data, which can be used to identify NLOS/multipath signals based on signal quality features \cite{li_machine_2023,garcia_crespillo_robust_2023,ali_explainable_2023}. A weighting scheme for precise point positioning (PPP) proposed by Li et al. \cite{li_efficient_2024} incorporates elevation angle, carrier-to-noise density ratio $C/N_0$, and the predicted NLOS probability from machine learning (ML) output. Among the popular ML algorithms, random forest achieves the highest classification accuracy \cite{li_efficient_2024}. A support vector machine (SVM) classifier that detects NLOS signals achieves approximately 90\% accuracy with static data and 86\% accuracy with dynamic data collected in downtown Tokyo \cite{ozeki_gnss_2022}. Crespillo et al. \cite{garcia_crespillo_robust_2023} developed a branched logistic regression algorithm employing multi-frequency GNSS features from different constellations independently. This algorithm achieves 84\% accuracy on static GPS dataset and 91.4\% accuracy on Galileo dataset \cite{garcia_crespillo_robust_2023}. Xu et al. \cite{xu_machine_2020} constructs a scoring system by incorporating SVM-based NLOS classifier with a shadow matching technique. The scoring system assigns scores to particles that are generated around the initial weighted least squares (WLS) estimate using Gaussian distribution \cite{xu_machine_2020}. The final position solutions are computed using particles whose scores exceed a predefined threshold \cite{xu_machine_2020}. Suzuki and Amano \cite{suzuki_nlos_2021} developed both SVM and neural network classifiers for NLOS signal detection utilizing geometric features of correlation function. Although the neural network demonstrates slightly better performance, its training time is 5 to 10 times longer than that of the SVM. Comparative studies in \cite{li_machine_2023} and \cite{koiloth_ml-based_2024} show that ensemble machine learning algorithms can achieve higher NLOS classification accuracy than other ML methods. Therefore, ensemble machine learning algorithms are employed in our work.

Following NLOS/multipath signal detection, excluding the degraded signals can adversely affect satellite geometry \cite{li_efficient_2024}. As a result, a WLS method is proposed to address the varying quality of pseudorange measurements \cite{teunissen_springer_2017}. WLS modifies the relative contribution of each signal through suitable weighting that reflects measurement quality \cite{tabatabaei_reliable_2017,ng_improved_2020,akram_gnss_2018,li_doppler-aided_2011}. Under the condition that weights correspond to the inverse of variance-covariance matrix of pseudorange errors, WLS estimation satisfies the best linear unbiased estimator (BLUE) criterion \cite{teunissen_springer_2017,li_doppler-aided_2011}. Traditional weighting schemes employ elevation angle and $C/N_0$ to construct variance estimation formulas \cite{li_combined_2022,ng_improved_2020,li_improvement_2009}. Machine learning approaches can capture the complex relationships between selected features and signal quality to improve variance estimation. An SVM-driven weighting approach for PPP was introduced by Lyu and Gao \cite{lyu_svm_2020}. The results demonstrate a 65.4\% improvement in positioning accuracy compared to traditional $C/N_0$-based weighting models \cite{lyu_svm_2020}. The random forest-based weighting approach developed by Li et al. \cite{li_machine_2023} outperforms other traditional weighting methods in terms of 3D mean error and standard deviation, and has been used also in \cite{lee-leib2025}.

Motivated by  machine learning NLOS/multipath detection, our work \cite{pin-thesis} considers a novel ensemble learning-based activation function framework for multi-constellation GNSS WLS positioning. The activation function framework is general in the sense that it can be used with other machine learning algorithms, beyond those used in this study. The ensemble learning models are trained to identify quality of signals based on multiple signal quality indicators. The activation functions are used to map the quality scores from machine learning output into weights. Multi-constellation GNSS effectively increases signal availability which is beneficial to the positioning accuracy especially in urban environments. An algorithm based on WLS  is employed to relatively increase the contribution of reliable signals  without degrading satellite geometry in the process of calculating  a position solution. The main contributions of our paper are three fold: (1) three ensemble learning techniques are employed to assess GNSS signal quality scores using indicators that reflect NLOS/multipath errors and satellite geometry; (2) with respect to our initial contribution \cite{lee-leib2025} the present paper introduces a novel activation function framework  and presents performance results with several functions  to identify the optimal one that minimizes the 3D positioning errors; (3) real-world datasets collected from Hong Kong and Tokyo are employed to validate the effectiveness of the proposed algorithm across single and multi-constellation configurations.

The rest of this paper is organized as follows: Section \ref{sec:methodology} provides an overview of the proposed algorithm, followed by an introduction of pseudorange model and WLS algorithm. The selected signal quality features are also discussed in this section. Section \ref{sec:ml} derives the machine learning algorithms and presents a selection of activation functions. Section \ref{sec:results} describes the experiment setup, and provides comprehensive analysis of results from both Hong Kong and Tokyo datasets. Section \ref{sec:conclusion} provides summary of research outcomes and prospective research directions.

\section{The Framework, Algorithm, and Models} \label{sec:methodology}

This section presents the concepts that lie the foundations of our framework, including features selection for Machine Learning applications  in satellite navigation.
\subsection{Overview}
\begin{figure}[h]
\centering
\includegraphics[width=2.5in]{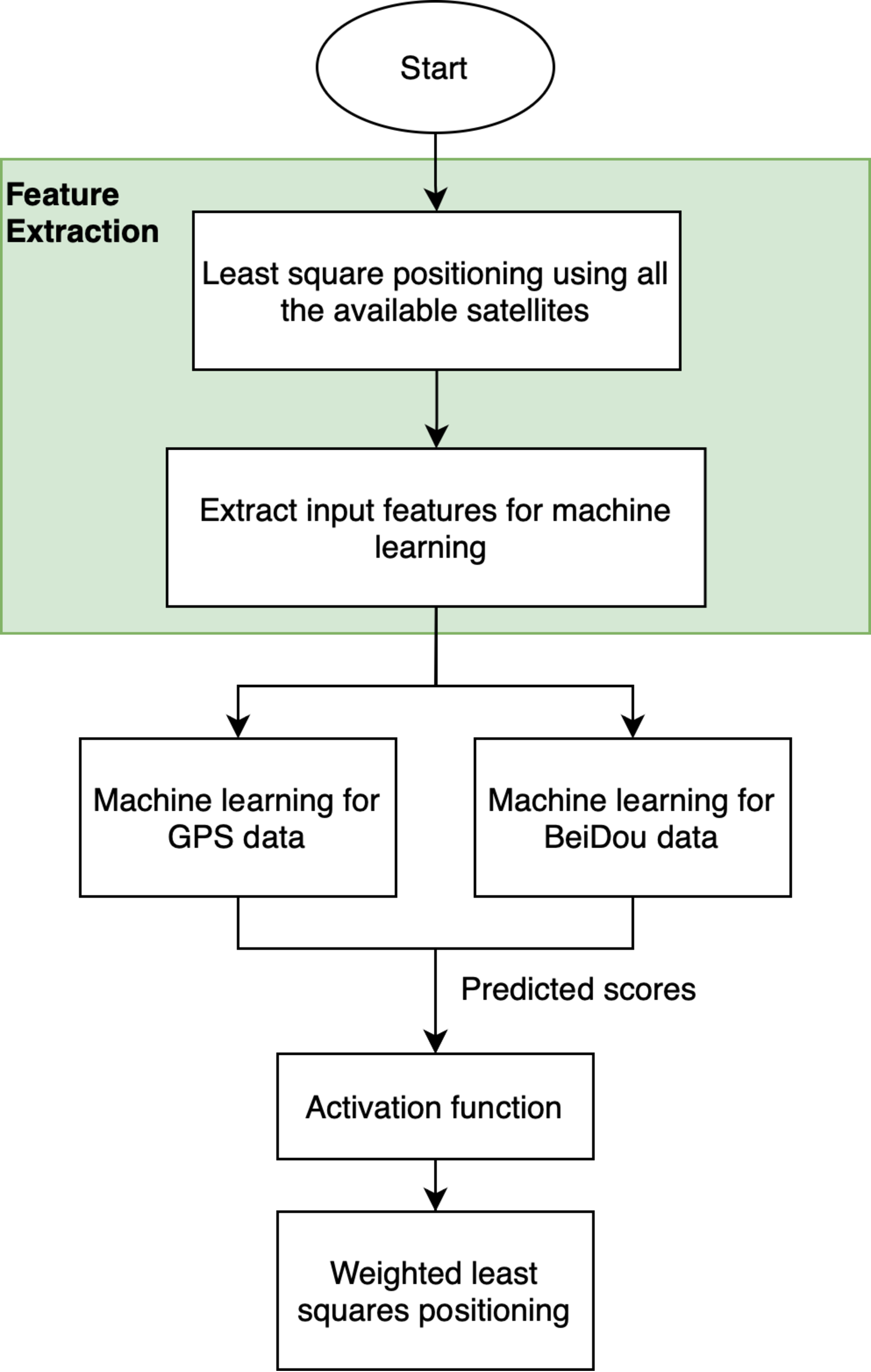}
\caption{Flowchart of the proposed algorithm.}
\label{fig:flowchart}
\end{figure}

For readers' convenience, in Figure \ref{fig:flowchart} we present the flowchart of the proposed algorithm \cite{pin-thesis}. Initially, ordinary least squares (OLS) positioning solutions are computed using all the available satellites. During this stage, signal quality features such as elevation angle, pseudorange residual, geometric dilution of precision (GDOP) contribution, pseudorange rate consistency, and estimated receiver clock offset are calculated and extracted. These features along with $C/N_0$ are then extracted to form machine learning training and testing data. In the training set, epochs with four or less signals are discarded to ensure the effectiveness of machine learning training process. Next, machine learning models are trained for GPS  and BeiDou data separately. The predicted scores from the machine learning output are then mapped into weights by the activation functions. Finally, these weights are employed in the WLS algorithm to calculate the final positioning solutions.

\subsection{Pseudorange Model}
The pseudorange $\rho$ \cite{groves_principles_2013} is modeled by:
\begin{equation}
    \rho_{i,j}^{model} = \sqrt{(x_{i,j}-x_u)^2+(y_{i,j}-y_u)^2+(z_{i,j}-z_u)^2}+\delta_\rho^{i} 
    \label{eqn:modeled_pseudorange}
\end{equation}
where $(x,y,z)$ is the 3-dimensional (3D) position in earth-centered earth-fixed (ECEF) coordinates, subscript $u$ indicates user, subscripts $(i,j)$ represent $j$-th satellite in $i$-th constellation and $\delta_\rho^{i}$ is the geometric distance equivalent of receiver clock offsets with respect to GNSS system time. The satellite position can be computed by using the parameters from GNSS navigation file. Given pseudorange measurements from $m$ GPS signals and $n$ BeiDou signals, the unknown parameters of the modeled pseudorange in \eqref{eqn:modeled_pseudorange} are represented by a 5 dimensional vector $\mathbf{x}_u=\begin{bmatrix}
        x_u & y_u & z_u & \delta_{\rho_u}^{\text{GPS}} & \delta_{\rho_u}^{\text{BD}}
    \end{bmatrix}^T$.
The modeled pseudorange in \eqref{eqn:modeled_pseudorange} is a non-linear function of user position coordinates $(x_u,y_u,z_u)$ and $\delta_\rho^{i}$. However it can be transformed into a set of  linear equations through a first-order Taylor expansion around an initial point $\mathbf{x}_0=\begin{bmatrix}
    x_0&y_0&z_0&\delta_{\rho_0}^{\text{GPS}}&\delta_{\rho_0}^{\text{BD}}
\end{bmatrix}^T$:
\begin{equation}
\begin{aligned}
    \rho^{model}(\mathbf{x}_u)&\approx
\rho^{model}(\mathbf{x}_0)+\frac{\partial\rho}{\partial x}(x_u-x_0) +\frac{\partial\rho}{\partial y}(y_u-y_0)+\frac{\partial\rho}{\partial z}(z_u-z_0)+\frac{\partial\rho}{\partial \delta_{\rho}^{\text{GPS}}}(\delta_{\rho_u}^{\text{GPS}}-\delta_{\rho_0}^{\text{GPS}})+\\
& \hspace*{115mm}\frac{\partial\rho}{\partial \delta_{\rho}^{\text{BD}}}(\delta_{\rho_u}^{\text{BD}}-\delta_{\rho_0}^{\text{BD}})\\
    &=\rho^{model}(\mathbf{x}_0)+\frac{\partial\rho}{\partial x}\Delta_x+\frac{\partial\rho}{\partial y}\Delta_y+\frac{\partial\rho}{\partial z}\Delta_z+\frac{\partial\rho}{\partial \delta_{\rho}^{\text{GPS}}}\Delta_{\delta_\rho^{\text{GPS}}}+\frac{\partial\rho}{\partial \delta_{\rho}^{\text{BD}}}\Delta_{\delta_\rho^{\text{BD}}}
\end{aligned}
    \label{eqn:taylor_expansion_1}
\end{equation}
where $\mathbf{\Delta_x}=\begin{bmatrix}\Delta_x&\Delta_y&\Delta_z&\delta_{\rho_u}^{\text{GPS}}&\delta_{\rho_u}^{\text{BD}}\end{bmatrix}^T=\mathbf{x}_u-\mathbf{x}_0$.

\subsection{Fundamentals of GNSS WLS Positioning}
The pseudorange measurement $\rho^{obs}$ from a GNSS receiver is subject to various errors, therefore it can be expressed as:
\begin{equation}
    \rho^{obs}_{i,j}=\rho^{model}_{i,j}+\epsilon_{i,j}
    \label{eqn:obs_pseudorange}
\end{equation}
where $\epsilon$ denotes the pseudorange measurement error and subscripts $(i,j)$ denote the $j$-th satellite in $i$-th constellation.  From \eqref{eqn:modeled_pseudorange} and \eqref{eqn:obs_pseudorange} we have for $m$ GPS  and $n$ BeiDou signals the linear equations for  residuals:
\begin{equation}
    \mathbf{\Delta}_{\boldsymbol{\rho}}=\mathbf{H}\mathbf{\Delta}_\mathbf{x}+\mathbf{\epsilon}
\end{equation}
where
$\mathbf{\Delta}_{\boldsymbol{\rho}}$ is the pseudorange residual vector:
\begin{equation}
    \mathbf{\Delta}_{\boldsymbol{\rho}}=\begin{bmatrix}
        \rho^{obs}_{\text{GPS},1}-\rho^{model}_{\text{GPS},1}(\mathbf{x}_0)\\
        \vdots\\
        \rho^{obs}_{\text{GPS},m}-\rho^{model}_{\text{GPS},m}(\mathbf{x}_0)\\
        \rho^{obs}_{\text{BD},1}-\rho^{model}_{\text{BD},1}(\mathbf{x}_0)\\
        \vdots\\
        \rho^{obs}_{\text{BD},n}-\rho^{model}_{\text{BD},n}(\mathbf{x}_0)
    \end{bmatrix}_{(m+n)\times1}
\end{equation}
 $\mathbf{H}$ is the measurement matrix:
\begin{equation}
    \mathbf{H}=\begin{bmatrix}
            \frac{\partial\rho_{\text{GPS},1}^{model}}{\partial x} & \frac{\partial\rho_{\text{GPS},1}^{model}}{\partial y} & \frac{\partial\rho_{\text{GPS},1}^{model}}{\partial z} & \frac{\partial\rho_{\text{GPS},1}^{model}}{\partial \delta_{\rho}^{\text{GPS}}} & \frac{\partial\rho_{\text{GPS},1}^{model}}{\partial \delta_{\rho}^{\text{BD}}}\\
            \vdots & \vdots & \vdots & \vdots & \vdots\\
            \frac{\partial\rho_{\text{GPS},m}^{model}}{\partial x} & \frac{\partial\rho_{\text{GPS},m}^{model}}{\partial y} & \frac{\partial\rho_{\text{GPS},m}^{model}}{\partial z} & \frac{\partial\rho_{\text{GPS},m}^{model}}{\partial \delta_{\rho}^{\text{GPS}}} & \frac{\partial\rho_{\text{GPS},m}^{model}}{\partial \delta_{\rho}^{\text{BD}}}\\
            \frac{\partial\rho_{\text{BD},1}^{model}}{\partial x} & \frac{\partial\rho_{\text{BD},1}^{model}}{\partial y} & \frac{\partial\rho_{\text{BD},1}^{model}}{\partial z} & \frac{\partial\rho_{\text{BD},1}^{model}}{\partial \delta_{\rho}^{\text{GPS}}} & \frac{\partial\rho_{\text{BD},1}^{model}}{\partial \delta_{\rho}^{\text{BD}}}\\
            \vdots & \vdots & \vdots & \vdots & \vdots\\
            \frac{\partial\rho_{\text{BD},n}^{model}}{\partial x} & \frac{\partial\rho_{\text{BD},n}^{model}}{\partial y} & \frac{\partial\rho_{\text{BD},n}^{model}}{\partial z} & \frac{\partial\rho_{\text{BD},n}^{model}}{\partial \delta_{\rho}^{\text{GPS}}} & \frac{\partial\rho_{\text{BD},n}^{model}}{\partial \delta_{\rho}^{\text{BD}}}
        \end{bmatrix}_{(m+n)\times5}
        \label{eqn:measurement_matrix}
\end{equation}
and
$\mathbf{\epsilon}$ is the pseudorange measurement error vector: $\mathbf{\epsilon}=\begin{bmatrix}
    \epsilon_{\text{GPS},1}&\dots&\epsilon_{\text{GPS},m}&\epsilon_{\text{BD},1}&\dots&\epsilon_{\text{BD},n}
\end{bmatrix}^T$.

The objective of WLS is to find the unknowns $\mathbf{\Delta}_\mathbf{x}$ that minimize the weighted sum of squared residuals: 
\begin{equation}
    \widehat{\mathbf{\Delta}_\mathbf{x}} = \arg \min_{\mathbf{x}_u}(\mathbf{\Delta}_{\boldsymbol{\rho}}-\mathbf{H}\mathbf{\Delta}_\mathbf{x})^T\boldsymbol{W}(\mathbf{\Delta}_{\boldsymbol{\rho}}-\mathbf{H}\mathbf{\Delta}_\mathbf{x})
    \label{eqn:wls_objective}
\end{equation}
where $\boldsymbol{W}$ is the weighting matrix \cite{teunissen_springer_2017}. The WLS problem in \eqref{eqn:wls_objective} is solved by:
\begin{equation}
    \widehat{\mathbf{\Delta}_\mathbf{x}}=(\mathbf{H}^T\boldsymbol{W}\mathbf{H})^{-1}\mathbf{H}^T\boldsymbol{W}\mathbf{\Delta}_{\boldsymbol{\rho}}
    \label{eqn:wls_solution}
\end{equation}

\subsection{Evaluation Metrics of GNSS Positioning}
Positioning performance is assessed through 3D errors for individual epochs and 3D root mean square error (RMSE) computed over all available epochs. The 3D positioning errors in the east-north-up (ENU) reference frame are computed as follows:
\begin{equation}
        Error_{3D}=\sqrt{(\hat{x}-x_{truth})^2+(\hat{y}-y_{truth})^2+(\hat{z}-z_{truth})^2}
        \label{eqn:error_3d}
\end{equation}
where $(\hat{x},\hat{y},\hat{z})$ and $(x_{truth},y_{truth},z_{truth})$ are estimated position and ground truth in east, north, and up direction, respectively. The 3D RMSE quantifies overall positioning accuracy across all epochs and is calculated as:
\begin{equation}
    RMSE_{3D} = \sqrt{\frac{1}{T}\sum_{t=1}^T(Error_{3D,t})^2}
    \label{eqn:rmse}
\end{equation}
where $T$ is the number of available epochs and the subscript $t$ indicates the $t$-th epoch.

\subsection{Feature Selection}
\subsubsection{Elevation Angle}
The elevation angle of a satellite is defined as the angle formed between the user’s local horizontal plane and the line-of-sight (LOS) vector directed toward the satellite \cite{groves_principles_2013}. Satellite signals from lower elevation angles are more susceptible to signal obstruction, reflection, and refraction caused by high-rise buildings. Consequently, the elevation angle is commonly used as an indicator for identifying NLOS  and multipath errors in pseudorange measurements \cite{ozeki_gnss_2022}.

\subsubsection{Carrier-to-noise Density}
$C/N_0$ is the ratio of carrier power to thermal noise power in 1Hz bandwidth \cite{teunissen_springer_2017}. Destructive multipath reception  and NLOS propagation typically cause substantial attenuation of $C/N_0$ \cite{zhang_site-specific_2018}. In contrast, constructive multipath signal can lead to an increase in $C/N_0$ \cite{zhang_site-specific_2018}. As a result, a low $C/N_0$ can indicate higher possibility of NLOS/multipath signal. Nevertheless, signals exhibiting high $C/N_0$ values could represent either high quality LOS signals or the results of constructive multipath.

\subsubsection{Pseudorange Residual}
Pseudorange residual is defined as the difference between pseudorange measurement and the modeled pseudorange using estimated user position and receiver clock bias. It is commonly used in Receiver Autonomous Integrity Monitoring (RAIM) to detect faulty pseudorange measurement. Pseudorange residual $R$ can be expressed as:
\begin{equation}
    R_{i,j} = \rho^{obs}_{i,j}-\rho_{i,j}^{model}(\widehat{\mathbf{x}}_u)
\end{equation}
where $\rho^{obs}$ is the pseudorange measurement that can be directly taken from GNSS observation (RINEX) file, $\rho^{model}$ is calculated from \eqref{eqn:modeled_pseudorange} using the estimated
position and receiver clock offset from all available satellites. A large pseudorange residual indicates a potential error in the pseudorange measurement.  

\subsubsection{GDOP Contribution}
GDOP serves as a quality indicator for satellite geometry, where a lower value denotes a better  geometry. The GDOP of $m+n$ satellites from two constellations is computed by:
\begin{equation}
    G_{m+n}=\sqrt{\operatorname{tr}((\mathbf{H}^T_{5\times (m+n)}\mathbf{H}_{5\times (m+n)})^{-1})}
\end{equation}
where $\operatorname{tr}(\cdot)$ denotes the matrix trace operator and $\mathbf{H}$ is the measurement matrix defined in \eqref{eqn:measurement_matrix}. The GDOP contribution of a specific satellite measures the impact of the  satellite signal due to satellite geometry \cite{li_modified_2012}. To obtain the GDOP contribution of satellite $k$, we first calculate the GDOP without the $k$-th satellite:
\begin{equation}
    G_{m+n-1}^k = \sqrt{\operatorname{tr}((\mathbf{H}^{(k)T}_{5\times(m+n-1)}\mathbf{H}^{(k)}_{5\times(m+n-1)})^{-1})}
\end{equation}
Subsequently, the GDOP contribution $\Delta_G^k$ of the $k$-th satellite is computed by:
\begin{equation}
    \Delta_G^k = G_{m+n-1}^k-G_{m+n}
\end{equation}
where $\Delta_G^k\geq0$ \cite{yarlagadda_gps_2000}. A large $\Delta_G^k$ indicates that excluding the $k$-th satellite would remarkably compromise satellite geometry. Hence, satellites with higher $\Delta_G^k$ values should receive greater weighting in WLS positioning computations.

\subsubsection{Pseudorange Rate Consistency}
Doppler-based measurements generally are more accurate than pseudorange measurements, as they are less sensitive to slowly changing error sources such as atmospheric delays and hardware biases \cite{teunissen_springer_2017}. This characteristic makes the pseudorange rate $\dot{\rho}$ computed from Doppler shift more reliable
\cite{borio_identifying_2016}: 
\begin{equation}
    \dot{\rho} = f_{Doppler}\cdot \frac{c}{f_{carrier}}
    \label{eqn:rho_doppler}
\end{equation}
where $f_{Doppler}$ denotes the observed Doppler shift frequency, $f_{carrier}$ represents the carrier frequency and $c$ is the speed of light. In contrast, the pseudorange rate calculated by the rate of change of pseudorange measurements can contain larger errors from more error sources. Therefore, pseudorange rate consistency $\epsilon_{\dot{\rho}}$ is adopted as a quality indicator and it can be calculated by:
\begin{equation}
    \epsilon_{\dot{\rho}}=\frac{(\rho_{t}^{obs}-\rho_{t-1}^{obs})}{T_e}-\dot{\rho}_t
\end{equation}
where $\rho^{obs}$ is the pseudorange measurement, the subscript $t$ indicates $t$-th epoch, $T_e$ is the time difference between two epochs and $\dot{\rho}$ is defined in \eqref{eqn:rho_doppler}. A large $\epsilon_{\dot{\rho}}$ suggest potential NLOS/multipath contamination in the pseudorange measurements \cite{hsu_gnss_2017}. 

\subsubsection{Estimated Receiver Clock Offset}
Estimated receiver clock offset $\hat{\delta}_\rho^i$ is calculated by:
\begin{equation}
    \hat{\delta}_\rho^i = \rho^{obs}_{i,j}-\hat{r}_{i,j}
\end{equation}
where  $\hat{r}=\sqrt{(x_{i,j}-\hat{x}_u)^2+(y_{i,j}-\hat{y}_u)^2+(z_{i,j}-\hat{z}_u)^2}$ is the geometric range between the satellite position $(x_{i,j},y_{i,j},z_{i,j})$ and estimated user position $(\hat{x}_u,\hat{y}_u,\hat{z}_u)$. 
The user position is estimated by LS positioning using all available satellites.
Assuming all the pseudorange measurements in a single epoch are accurate, the receiver clock offsets derived from all satellites $j=1,\dots,n$ in $i$-th constellation should be identical. Therefore, inconsistent estimated receiver clock offset for individual signals reveal possible NLOS/multipath degradation of the pseudorange measurements.

\subsection{Feature Normalization}
To distinguish the relative quality of each satellite signal in an epoch, Z-score standardization is performed on an epoch-by-epoch basis as follows:
\begin{equation}
    \tilde{d}_{j,t}=\frac{d_{j,t}-\bar{d}_t}{\sigma_t}
\end{equation}
where $d_{j,t}$ denotes feature of $j$-th signal in epoch $t$, $\tilde{d}$ is normalized feature, $\bar{d}_t$ and $\sigma_t$ are the mean value and standard deviation (STD) of the feature in epoch $t$, respectively. This normalization process eliminates the effect caused by different numerical scales of each feature and enhances machine learning accuracy. Furthermore, the absolute values are taken for normalized pseudorange residual, pseudorange rate consistency, and estimated receiver clock offset to quantify the magnitude of deviations from their respective means:
\begin{equation}
    \tilde{d}_{j,t} = \left| \frac{d_{j,t}-\bar{d}_t}{\sigma_t} \right|
\end{equation}
	
\section{Machine Learning Activation Function Framework} \label{sec:ml}
While deep learning approaches are able to learn complex patterns from raw features, they typically demand extensive training datasets and substantial computational resources to attain satisfactory performance \cite{goodfellow_deep_2016}. In comparison, given limited amount of data,  simpler structures of traditional machine learning can achieve comparable performance using less computation power. Furthermore, studies show that ensemble learning algorithms outperform other machine learning methods in NLOS classification tasks \cite{li_machine_2023, koiloth_ml-based_2024}. Hence, this study employs ensemble-based machine learning algorithms to evaluate signal quality.

\subsection{Data Labeling}
Data labeling assigns informative labels to features, which enables supervised machine learning algorithms to learn the relationship between input features and the target output. In this study, a label is assigned according to the best set of satellites  that produces the minimum 3D positioning errors among all the subset combinations using LS positioning. Satellite signals included in this best set are assigned a label of 1 and the remaining signals are labeled as 0. Such labelled data sets are essential for the training process.

\subsection{Machine Learning Algorithms}
\subsubsection{Bagging}
Bootstrap aggregating, commonly known as bagging, involves three stages: bootstrapping, training, and aggregation \cite{kumar_ensemble_2020}. Bootstrap sampling forms multiple subsets by randomly selecting data points with replacement from the original data \cite{bonaccorso_mastering_2020}. Each data point includes feature vectors with associated labels. Subsequently, each subset is employed to train a weak learner in parallel. A weak learner is a simple model that has slightly better accuracy compared to random guessing. During aggregation, predictions from all weak learners are combined through methods such as majority voting or soft voting \cite{kumar_ensemble_2020}.

\begin{figure}[h!]
\centering
\includegraphics[width=2in]{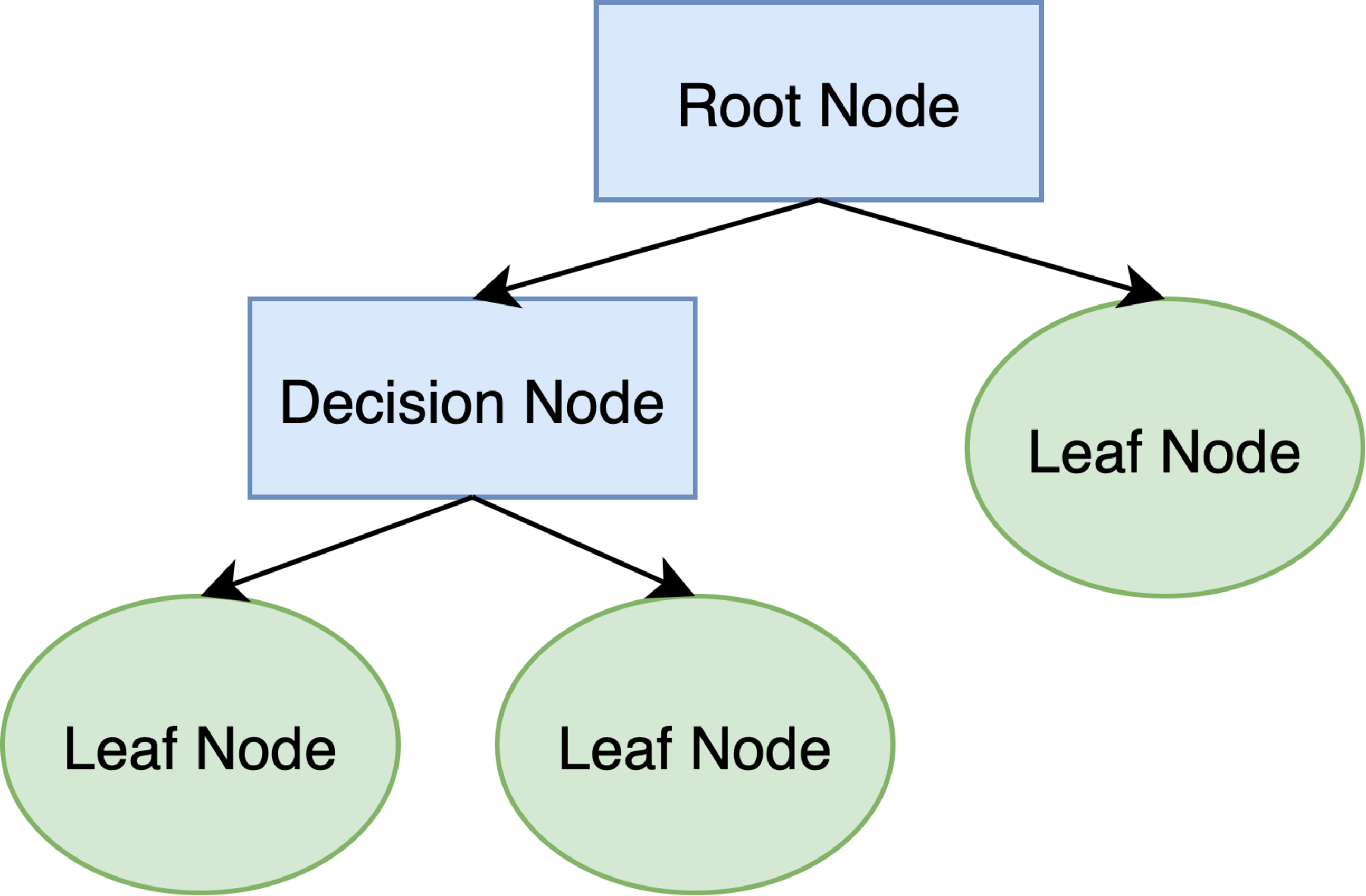}
\caption{Decision Tree.}
\label{fig:decision_tree}
\end{figure}

Random forest is one of the bagging algorithms that employs decision trees as weak learners. Figure \ref{fig:decision_tree} illustrates the hierarchical structure and elements of a decision tree. Each decision tree starts with a root node, which contains all the data to be classified. The root node and subsequent decision nodes continuously split the data using specific criteria until a stopping condition is met \cite{alpaydin_introduction_2020}. The branch connections represent decision pathways linking decision nodes to their child nodes. The leaf nodes are the final classification results at the bottom of the decision tree \cite{zhou_machine_2022}. The best splitting criterion maximizes the purity of resulting child nodes, where purity is measured using metrics such as Gini impurity and entropy. For $K$ classes classification, Gini impurity $H_{gini}$ is computed by:
\begin{align}
    \begin{split}
         H_{gini}&= \sum_{k=1}^Kp_k(1-p_k)
    \end{split}
\end{align}
and the entropy $H_{entropy}$ is calculated by:
\begin{equation}
    H_{entropy}= -\sum_{k=1}^Kp_klog(p_k)
\end{equation}
where $p_k$ is the proportion of samples that belong to $k$-th class in a node. A low Gini impurity or entropy indicates an effective split, as it reflects that the majority of data points inside each child node belong to a single class. Instead of considering all the features, random forest algorithm randomly selects a subset of features and decides a best split only using features within this subset. This approach enhances variability of base learners which provides  good generalization capabilities on real-world datasets \cite{zhou_machine_2022}.

\subsubsection{Boosting}
Unlike bagging algorithms that train independent weak learners in parallel, boosting approaches train weak learners sequentially where each successive weak learner focuses more on misclassified instances from previous weak learner \cite{kumar_ensemble_2020,zhou_ensemble_2012}.

Adaptive boosting, abbreviated as AdaBoost, is a boosting algorithm developed by Freund and Schapire \cite{freund_decision-theoretic_1997}. The algorithm trains individual weak learners using identical  data sets but applies different sample weights. Consider training data $\boldsymbol{S}$ with $N$ data points
\begin{equation}
    \boldsymbol{S} = \{(\boldsymbol{d}_1,l_1),(\boldsymbol{d}_2,l_2),...,(\boldsymbol{d}_N,l_N)\}
    \label{eqn:ml_training_set}
\end{equation}
where $\boldsymbol{d}$ denotes feature vector and $l$ is the corresponding label. The first weak learner $\phi_1$ is trained on training data $\mathbf{S}$ with uniform weights $w_n^1=\frac{1}{N}$. The error rate $\epsilon$ of $m$-th weak learner is calculated as the weighted proportion of misclassified instances:
\begin{equation}
    \epsilon^m = \frac{\sum_{n:\phi^m(\boldsymbol{d}_n)\neq l_n} w_n^m}{\sum_{n=1}^Nw_n^m}
    \label{ada_error_rate}
\end{equation}
where $m=1,2,...M$ indicates $m$-th weak learner. The weight of $m$-th weak learner $\alpha^m$ is then calculated based on its error rate to reflect its classification accuracy:
\begin{equation}
    \alpha^m = \eta \cdot log(\frac{1-\epsilon^m}{\epsilon^m})
    \label{ada_wl_weight}
\end{equation}
where $\eta$ is the learning rate specified by the user. To emphasize the errors, the sample weights of misclassified instances are increased while the weights of correct classifications are decreased according to:
\begin{equation}
    w_n^{m+1}=
    \begin{cases}
        \frac{w_n^m exp(-\alpha^m)}{Z^m},\quad \text{if }\phi^m(\boldsymbol{d}_n)= l_n \\
        \frac{w_n^m exp(\alpha^m)}{Z^m},\quad \text{if }\phi^m(\boldsymbol{d}_n)\neq l_n
    \end{cases}
    \label{ada_smaple_weight}
\end{equation}
where $Z^m$ is a normalization constant to ensure the sum of sample weights equals 1. The subsequent weak learner is trained on the training data $\mathbf{S}$ with the updated weights. After repeating this process for $M$ weak learners, the AdaBoost model is constructed by combining all weak learners and their weights:
\begin{equation}
    \Phi(\boldsymbol{d})=sign(\Sigma_{m=1}^M\alpha^m\phi^m(\boldsymbol{d}))
\end{equation}

Gradient boosting builds a strong classifier by sequentially fitting weak learners to the pseudo‐residuals which are negative gradients of the loss function \cite{friedman_greedy_2001}. Given training data in \eqref{eqn:ml_training_set}, the objective of gradient boosting is to construct a strong learner $\Phi(\boldsymbol{d})$ that minimizes the total loss over all the training samples:
\begin{equation}
    \hat{\Phi}(\boldsymbol{d}) = \arg \min_{\Phi(\boldsymbol{d})}\sum_{n=1}^N\Psi(l_n,\Phi(\boldsymbol{d}_n))
\end{equation}
where $\Psi(\cdot)$ is a loss function, $N$ is the number of samples and subscript $n$ indicates $n$-th sample. The log loss $\Psi(l_n,\Phi(\boldsymbol{d}_n))$ is commonly chosen as loss function for binary classification and it is developed from negative log likelihood:
\begin{align}
    \begin{split}
        \Psi(l_n,\Phi(\boldsymbol{d}_n))
        &=-\log L(l_n,\Phi(\boldsymbol{d}_n))=-\log (\hat{P}(l_n=1|\boldsymbol{d}_n)^{l_n} \cdot \hat{P}(l_n=0|\boldsymbol{d}_n)^{1-l_n})\\
        &=-l_n\cdot\beta+\log(1+e^\beta)
    \end{split}
\end{align}
where $\beta=\Phi(\boldsymbol{d}_n)$ is the log-odds ratio:
\begin{equation}
    \beta=\log\frac{\hat{P}(l_n=1|\boldsymbol{d}_n)}{\hat{P}(l_n=0|\boldsymbol{d}_n)}.
\end{equation}
Gradient boosting begins with an initial prediction model $\Phi_0(\boldsymbol{d})$ set to the log-odds ratio of the training samples:
\begin{equation}
    \Phi_0(\boldsymbol{d}) = \log \frac{\sum_{n=1}^NI(l_n=1)}{N-\sum_{n=1}^NI(l_n=1)}
    \label{eqn:gb_initial_model}
\end{equation}
where $I(\cdot)$ is the indicator function. For iteration $m=1$ to $M$, the pseudo-residual $\tilde{l}_{n}^m$ is obtained by computing the negative gradient of the log-loss function:
\begin{align}
    \begin{split}
        \tilde{l}_{n}^m&=-\left[\frac{\partial \Psi(l_n,\Phi(\boldsymbol{d}_n))}{\partial \Phi(\boldsymbol{d}_n)}\right]_{\Phi(\boldsymbol{d}_n)=\Phi^{m-1}(\boldsymbol{d}_n)}
        =-\left[\frac{\partial (-l_n\cdot \Phi(\boldsymbol{d}_n)+\log(1+e^{\Phi(\boldsymbol{d}_n)}))}{\partial \Phi(\boldsymbol{d}_n)}\right]_{\Phi(\boldsymbol{d}_n)=\Phi^{m-1}(\boldsymbol{d}_n)}\\
        &=-\left[-l_n+\frac{e^{\Phi(\boldsymbol{d}_n)}}{1+e^{\Phi(\boldsymbol{d}_n)}}\right]_{\Phi(\boldsymbol{d}_n)=\Phi^{m-1}(\boldsymbol{d}_n)}
        =l_n-\frac{e^{\Phi^{m-1}(\boldsymbol{d}_n)}}{1+e^{\Phi^{m-1}(\boldsymbol{d}_n)}}
    \end{split}
\end{align}
Since $\frac{e^{\Phi^{m-1}(\boldsymbol{d}_n)}}{1+e^{\Phi^{m-1}(\boldsymbol{d}_n)}}=\hat{P}(l_n=1|\boldsymbol{d}_n)$, the pseudo-residual $\tilde{l}_{n}^m$ represents the discrepancy between the true label and the predicted probability. An updated training dataset is formed by using the pseudo-residuals as labels:
\begin{equation}
    \boldsymbol{S}^m=\{(\boldsymbol{d}_1,\tilde{l}_{1}^m),(\boldsymbol{d}_2,\tilde{l}_{2}^m),\dots,(\boldsymbol{d}_N,\tilde{l}_{N}^m)\}
\end{equation}
Using the updated training set $\mathbf{S}^m$, a weak learner $\phi^m(\boldsymbol{d},\boldsymbol{\theta}^m)$ is trained by solving the least squares problem:
\begin{equation}
    \boldsymbol{\theta}^m = \arg \min_{\boldsymbol{\boldsymbol{\theta}}}\sum_{n=1}^N \left( \tilde{l}_{n}^m - \phi^m(\boldsymbol{d}_n,\boldsymbol{\theta}^m)\right)^2
\end{equation}
where $\boldsymbol{\theta}^m$ is the parameters of $m$-th weak learner. Finally, the strong learner is updated by adding the trained weak learner:
\begin{equation}
    \Phi^m(\boldsymbol{d}) = \Phi^{m-1}(\boldsymbol{d})+\eta \cdot\phi^m(\boldsymbol{d},\boldsymbol{\theta}^m)
\end{equation}
where $\eta$ is the learning rate specified by the user. After executing $M$ iterations, the final prediction model is given by:
\begin{equation}
    \Phi^M(\boldsymbol{d}) = \Phi^0(\boldsymbol{d}) + \sum_{m=1}^M \eta \cdot \phi^m(\boldsymbol{d},\boldsymbol{\theta}^m)
\end{equation}

\subsubsection{Evaluation Metrics of Machine Learning}
To evaluate the machine learning performance, we use 0.5 as a threshold. When the predicted scores is larger than 0.5 it is treated as label 1; otherwise, label 0. The accuracy is calculated by:
\begin{equation}
    \text{Accuracy} = \frac{\text{Number of Correct Predictions}}{\text{Total Number of Predictions}}.
\end{equation}
When calculating the testing accuracy, we omit the epochs with four or less signals since all the available signals are used to produce positioning results. 

\subsection{Activation Function}
\begin{figure}[h!]
\centering
\includegraphics[width=2.5in]{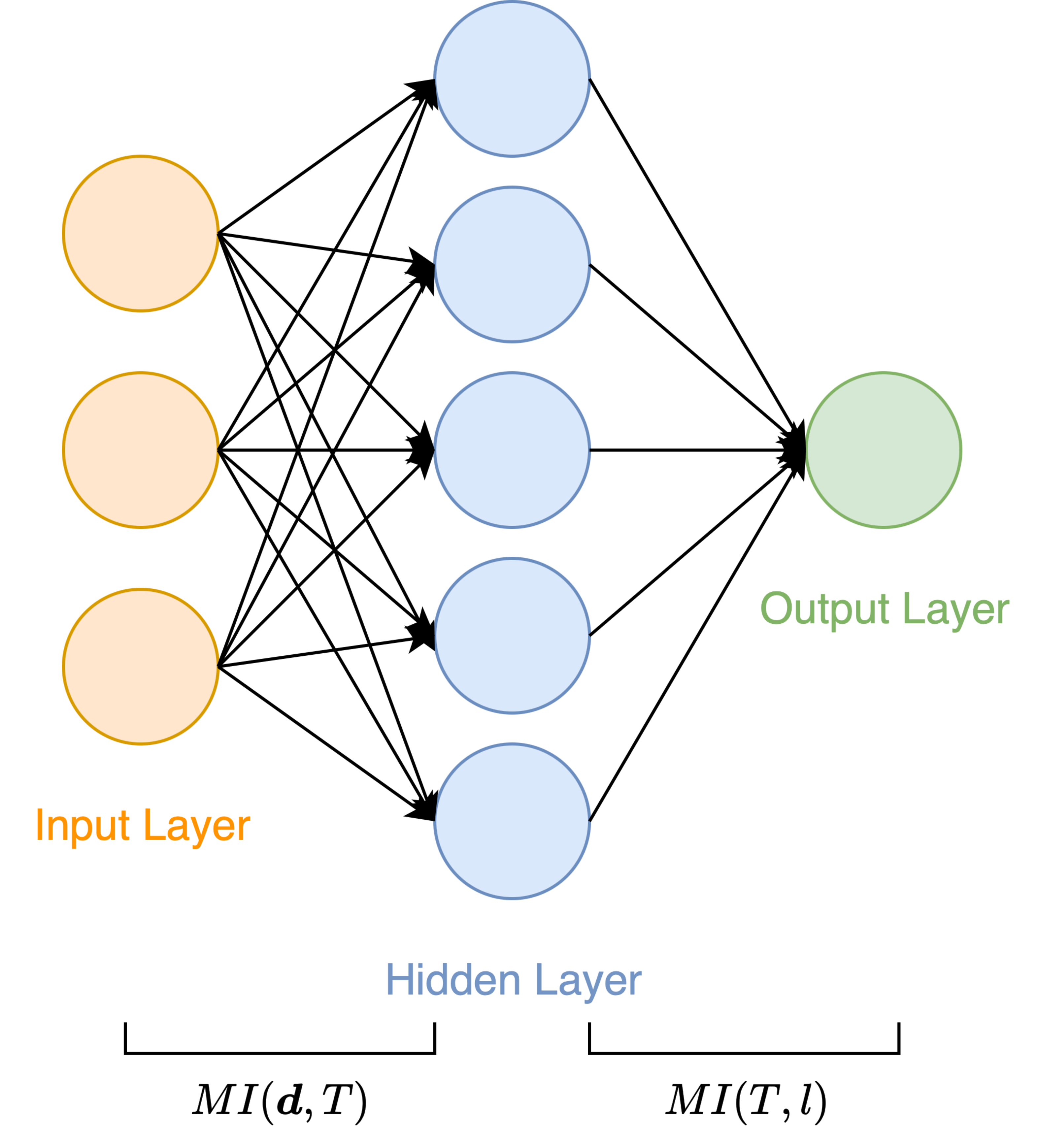}
\caption{Hidden Layer.}
\label{fig:hidden_layer}
\end{figure}

Inspired by  neural networks (NN), activation functions are employed to convert quality scores from machine learning output into appropriate weights for least squares positioning. Each layer of an NN represents a summary statistics of input where only part of the input information that maximizes the information retained about the target output is preserved \cite{saxe_information_2019}. Figure \ref{fig:hidden_layer} illustrates a hidden layer. The mutual information (MI) between hidden layer $T$ and input $\boldsymbol{d}$ is represented as $MI(\boldsymbol{d},T)$ and MI between hidden layer and output $l$ is denoted by $MI(T,l)$. An information plane is formed by $MI(\boldsymbol{d},T)$ and $MI(T,l)$ where the fitting phase and compression phase are observed during the training process \cite{saxe_information_2019}. Throughout the fitting phase, both $MI(\boldsymbol{d},T)$ and $MI(T,l)$ increase, suggesting that the hidden layer is learning to associate input features with corresponding output labels \cite{saxe_information_2019}. In the compression stage, the hidden layer discards the irrelevant input information which decreases $MI(\boldsymbol{d},T)$. A subsequent compression phase after fitting phase can be observed in the information plane when applying double-sided saturating functions such as tanh and sigmoid \cite{saxe_information_2019}. The compression phase starts when the activation function enters their saturation region. For linear functions or single-sided saturating functions such as rectified linear unit (ReLU), fitting and compression phases occur simultaneously and only the task-irrelevant information is compressed. Our approach leverages the compression properties of activation functions to suppress the noise or task-irrelevant information in machine learning outputs.

Furthermore, a NN relies on activation functions to introduce nonlinearities into the algorithm \cite{dubey_activation_2022}. The selection of appropriate activation functions significantly impacts the training process and performance. Non-linear activation functions are usually preferred over linear ones in practical scenarios \cite{sharma_activation_2020}. The saturation characteristics of certain activation functions assign zero weights to low-quality signals and uniform weights to high-quality signals, while moderately weighting intermediate-quality signals to preserve beneficial satellite geometry and improve the accuracy of the LS solution. In this study, the input to the activation function is the predicted probability from the machine learning, ranging from $[0,1]$ while the corresponding outputs are maintained within the same $[0,1]$ interval.

\begin{figure}[h!]
\centering
\makebox[\hsize]{%
        \includegraphics[width=8in]{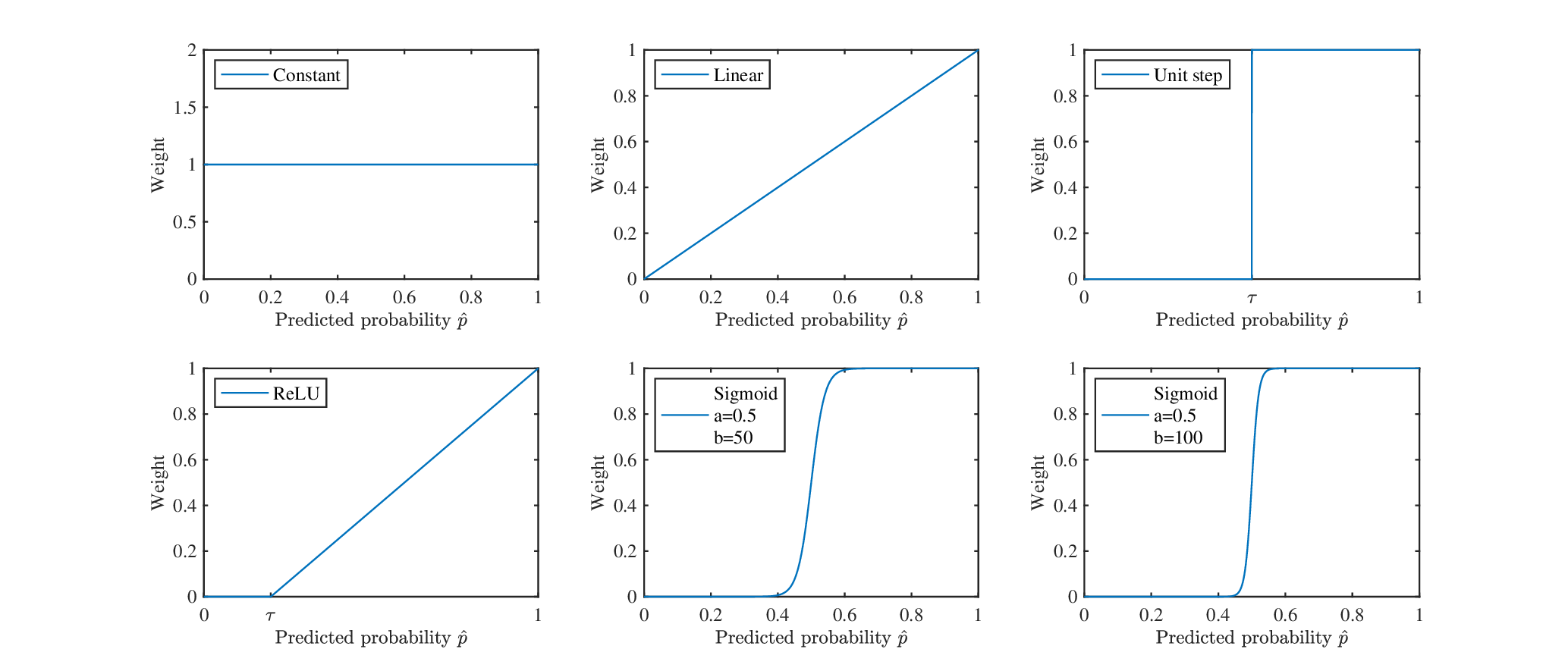}%
    }
\caption{Activation Functions.}
\label{fig:activation_function}
\end{figure}

Several activation functions are selected to assess their effects on weight design in the WLS framework. Figure \ref{fig:activation_function} illustrates the selected activation functions. The constant function that assigns identical weights to all measurement serves as a baseline. It is expressed as $f_{const}(\hat{p})=1$ where $\hat{p}$ is the predicted probability from machine learning output. The WLS problem turns into OLS problem in this case. The linear function directly translates the machine learning output scores into weights. It is expressed as $f_{linear}(\hat{p})=\hat{p}$. The unit-step function implements a hard threshold by selecting only a subset of satellites to produce an OLS solution. The predicted scores below the threshold are assigned zero weights and the remaining signals with predicted scores above the threshold are given unit weights. The unit step function $f_{unit}(\cdot)$ can be expressed as:
\begin{equation}
    f_{unit}(\hat{p})=
    \begin{cases}
        1,\quad \hat{p}\geq\tau\\
        0,\quad \hat{p}<\tau
    \end{cases}
\end{equation}
where $\tau\in[0,1]$ is the threshold. The threshold is set to the mean predicted scores of all available signals in a certain epoch. If the number of signals with unit weight is insufficient to generate LS results, the threshold is lowered by 0.05. ReLU activation function assigns zero weights to low-quality measurements whose predicted scores are below threshold $\tau$ and linearly maps the remaining scores to the range $[0,1]$. It is defined as:
\begin{equation}
    ReLU(\hat{p})=
    \begin{cases}
        \frac{\hat{p}-\tau}{1-\tau},\quad \hat{p}\geq\tau\\
        0,\quad \hat{p}<\tau
    \end{cases}
\end{equation}
where $\tau\in[0,1]$. In this study, $\tau$ is selected as the minimum predicted scores among all available signals in current epoch. The  sigmoid function makes soft decision that lowers weights for poor-quality signals and increases weights for high-quality ones. It is defined as:
\begin{equation}
    sigmoid(\hat{p};a,b)=\frac{1}{1+e^{-b(\hat{p}-a)}}.
    \label{eqn:sigmoid}
\end{equation}
where $a$ and $b$ controls the horizontal shift and steepness of the curve, respectively. When the steepness is infinitely large, the sigmoid function becomes a step function. To optimize the performance of the sigmoid function for various cases, we consider different values of shaping parameter $b$ in the positioning results while parameter $a$ is set to mean predicted scores of all available signals in the current epoch. 

\section{Experimental Results} \label{sec:results}

This section presents results obtained when applying our framework to real life measurements.

\subsection{Experimental Setup}
This study employs the UrbanNav open-source dataset, which provides GNSS observation data and ground truth information for datasets collected in Hong Kong and Tokyo urban areas \cite{hsu_hong_2023}. The GNSS observation data are written in Receiver Independent Exchange (RINEX) format. The ground truth data are written in txt files for Hong Kong data and csv files for Tokyo data. GPS L1 C/A and BeiDou B1I signals are used in this study. An elevation mask that excludes signals with elevation angle below 15$^\circ$ is applied to avoid low-quality signals. 

\subsubsection{Hong Kong Dataset}
For Hong Kong dataset, the GNSS observation data were recorded using a low-cost u-blox EVK-M8T receiver with 1 Hz output frequency. Ground truth data were collected by NovAtel SPAN-CPT receiver at an output frequency of 1 Hz with post-processing conducted by Inertial Explorer software \cite{hsu_hong_2023}. 

The Hong Kong dataset consists of data collected from three locations with different urbanization levels: medium urban in Tsim Sha Tsui, deep urban in Whampoa and harsh urban in Mong Kok \cite{hsu_hong_2023}. Medium urban data include 768 epochs, deep urban data consist of 1518 epochs and harsh urban data contain 2312 epochs. Training set for machine learning includes deep urban and harsh urban data while medium urban data serve as testing set. Figure \ref{subfig:medium_urban_GT} depicts the ground truth trajectories of medium urban data where the vehicle started at epoch 1, following a clockwise route for two loops and ended at epoch 768. 

\subsubsection{Tokyo Dataset}
For Tokyo dataset, GNSS observation data were collected by a u-blox ZED-F9P sensor at a 5Hz output frequency. Ground truth data were collected by Applanix POS LV620 at an output frequency of 10Hz. To maintain consistency with the Hong Kong dataset, both datasets were downsampled to 1 Hz by extracting measurements at integer seconds.

\begin{figure*}[!t]
\centering
\subfloat[]{\includegraphics[width=2.5in]{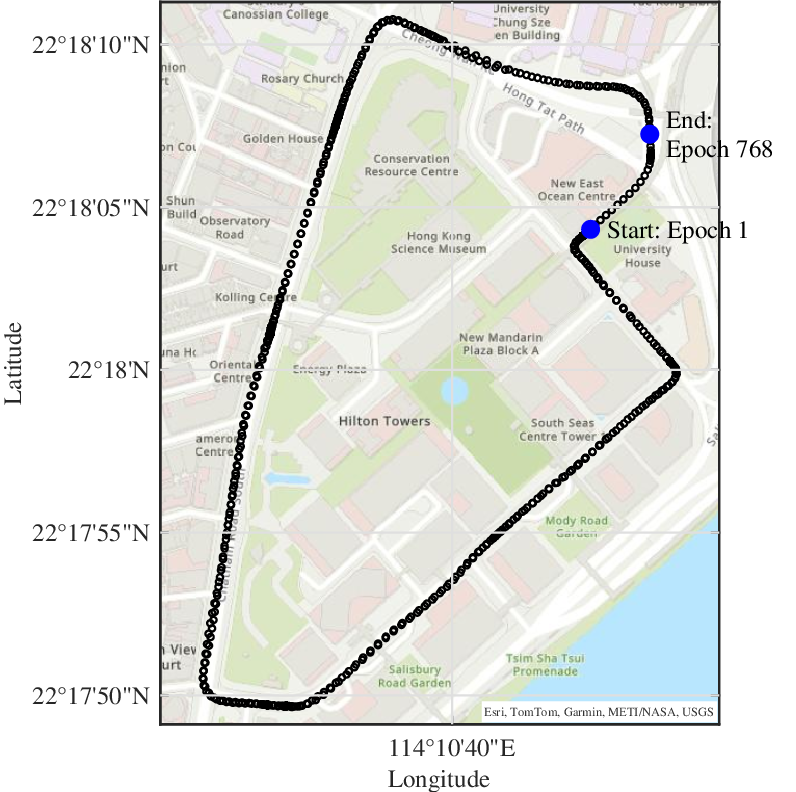}%
\label{subfig:medium_urban_GT}}
\hfil
\subfloat[]{\includegraphics[width=2.5in]{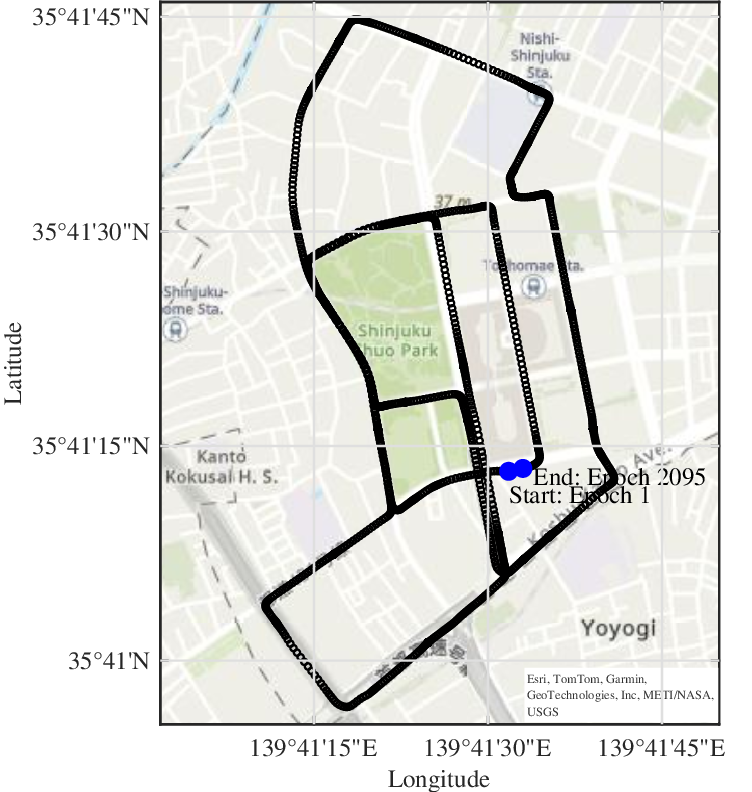}%
\label{subfig:shinjuku_GT}}
\caption{Ground truth of testing data (a) medium urban data from Hong Kong (b) Shinjuku data from Tokyo.}
\label{fig:GT}
\end{figure*}

The Tokyo experimental dataset comprises data from two urban locations: Shinjuku and Odaiba. Shinjuku data are collected in the dense urban environment whereas Odaiba data are collected in the suburban area with more open spaces and lower buildings. Odaiba data include 1241 epochs and Shinjuku data contain 2095 epochs. Figure \ref{subfig:shinjuku_GT} shows the ground truth of Shinjuku data where the vehicle follows clockwise route to circle around the area and the park from epoch 1 and stops at epoch 2095. Since Hong Kong dataset has comparable urbanization level as Shinjuku data, we use Shinjuku data as testing set and Hong Kong data as training set to verify the applicability of the proposed algorithm in different locations. In addition, we compare its performance with the case where machine learning models are trained on Odaiba data.

\subsection{Calibration using the CSRS-PPP Tool}
\begin{figure}
    \centering
    \includegraphics[width=0.5\linewidth]{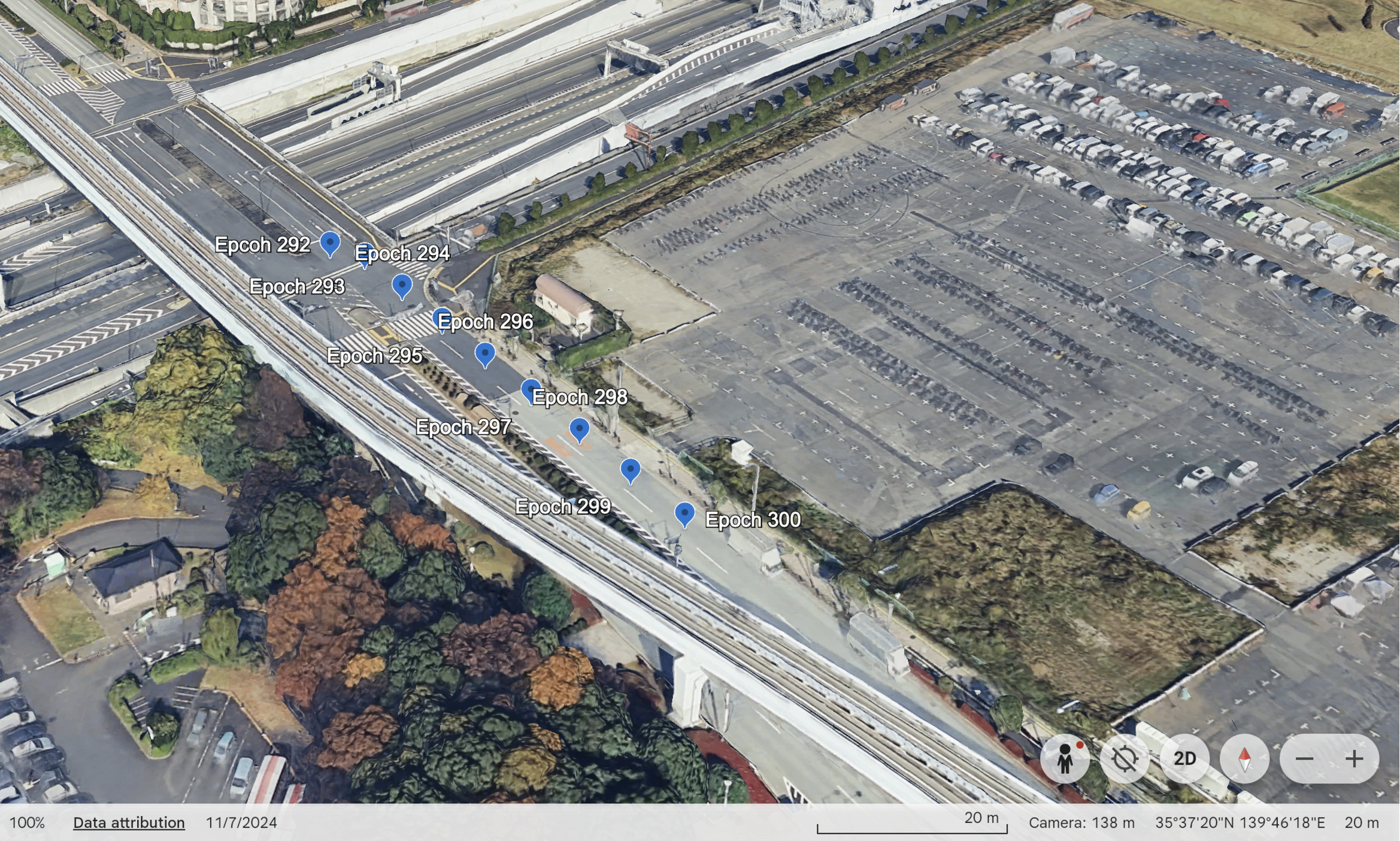}
    \caption{Google Earth 3D view of the open sky  ground truth trajectory for the Odaiba dataset epochs 291–300.}
    \label{fig:epoch_291_300}
\end{figure}

Firstly, we check the correctness of our  implemented baseline LS algorithm for positioning  using all available GPS+BeiDou satellites in an  open-sky  environment where machine learning is not used, since all available signals are not distorted. Such positioning results are compared with  solutions obtained from the Canadian Spatial Reference System Precise Point Positioning (CSRS-PPP) tool \cite{natural_resources_canada_canadian_nodate} when using the same data base. The CSRS-PPP service offers high-accuracy positioning solutions which we use  as a reference for comparison.. Figure \ref{fig:epoch_291_300} illustrates the ground truth trajectory for epochs 291–300 of the Odaiba data set corresponding to  an open-sky environment. Table \ref{tbl:csrs_ppp_compare} presents a comparison of the 3D positioning errors obtained when using our LS algorithm employing a total of 14 satellites (6 from GPS and 8 from BeiDou) and the CSRS-PPP solutions. The differences between LS and CSRS-PPP results are around 30 (cm). We see that our positioning  LS algorithm provides results that are consistent with those from the high-precision CSRS-PPP service in an open-sky environment. 
When using the CSRS-PPP tool on data obtained from non-open sky environments, where satellite signals may suffer severe propagation distortion, it is noticed that 43.37 \% of the epochs were rejected. This indicates that the CSRS-PPP tool is not efficient in handling densily built urban environments. These are precisely the environments where we envisage machine learning approaches to be useful.

\begin{table}[]
\centering
\caption{The 3D positioning errors in meters for epochs 291–300 under open-sky  using the LS algorithm with 14 GPS+BeiDou satellites and the CSRS-PPP tool.}
\begin{tabular}{|c|c|c|}
\hline
Epoch & LS 3D Error & CSRS-PPP 3D Error \\ \hline
291   & 2.1564   & 1.8546        \\ \hline
292   & 2.1754  & 1.8475         \\ \hline
293   & 2.1372  & 1.8582        \\ \hline
294   & 2.1493  & 1.8756        \\ \hline
295   & 2.1850  & 1.8904        \\ \hline
296   & 2.1895  & 1.9019        \\ \hline
297   & 2.2034  & 1.9134        \\ \hline
298   & 2.2320  & 1.8962        \\ \hline
299   & 2.3440   & 1.8955        \\ \hline
300   & 2.4671  & 1.8837        \\ \hline
\end{tabular}
\label{tbl:csrs_ppp_compare}
\end{table}

\subsection{Results of Hong Kong Dataset}
\subsubsection{Machine Learning Results}
The machine learning results for GPS data are summarized in Table \ref{tbl:hk_gps_ml}. The testing accuracies are $8.66\%$ and $8.23\%$ lower than training accuracies for random forest and gradient boosting algorithms, respectively. These results show that the machine learning models are affected by various distortions in the training data, and can lead to low generalization ability to unseen data in the testing process. In contrast, despite AdaBoost's lower training accuracy compared to random forest and gradient boosting approaches, its testing performance aligns closely with its training results and delivers the best testing accuracy among all the algorithms evaluated in this work.

For BeiDou data, the machine learning results are presented in Table \ref{tbl:hk_bd_ml}. Similar to the results of GPS data, random forest and gradient boosting suffer from overfitting with a significant drop of $6.96\%$ and $5.75\%$ from training to testing accuracies, respectively. AdaBoost maintains the most stable accuracies between training and testing results and achieves the highest testing accuracy at 70.43\%. 

Machine learning performance for GPS data is better than that for BeiDou data which might be reflected on  positioning accuracy. Since random forest and AdaBoost achieve comparable testing accuracies, the following section will consider positioning results of these two algorithms. We will start  with an analysis of activation functions, since they significantly affect positioning accuracy.

\begin{table}[!t]
\caption{Machine learning results for GPS with  Hong Kong Dataset.}
\label{tbl:hk_gps_ml}
\centering
\begin{tabular}{|c|c|c|c|}
\hline
                  & Random Forest & AdaBoost & Gradient Boosting \\ \hline
Training Accuracy &  0.8308  &    0.7603     &   0.8086                \\ \hline
Testing Accuracy  & 0.7442   &     0.7669    &  0.7263                 \\ \hline
\end{tabular}
\end{table}

\begin{table}[!t]
\caption{Machine learning results for BeiDou with Hong Kong Dataset.}
\label{tbl:hk_bd_ml}
\centering
\begin{tabular}{|c|c|c|c|}
\hline
                  & Random Forest & AdaBoost & Gradient Boosting \\ \hline
Training Accuracy & 0.7595   &   0.7133            &    0.7446               \\ \hline
Testing Accuracy  & 0.7020   &   0.7043            &    0.6750              \\ \hline
\end{tabular}
\end{table}

\subsubsection{Activation Function Analysis}
\begin{figure*}[!htbp]
\centering
\vspace*{-4mm}
\subfloat[]{\includegraphics[width=3.1in]{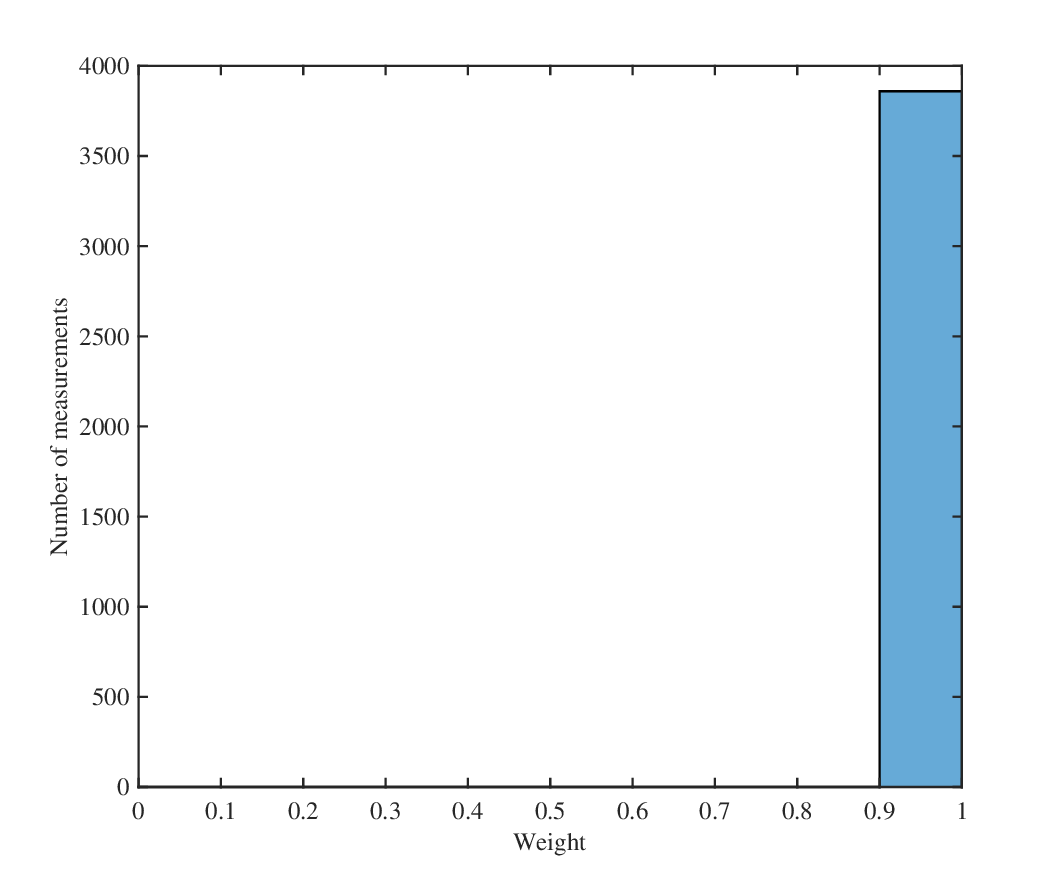}%
\label{subfig:constant}}
\hfil
\vspace*{-4mm}
\subfloat[]{\includegraphics[width=3.1in]{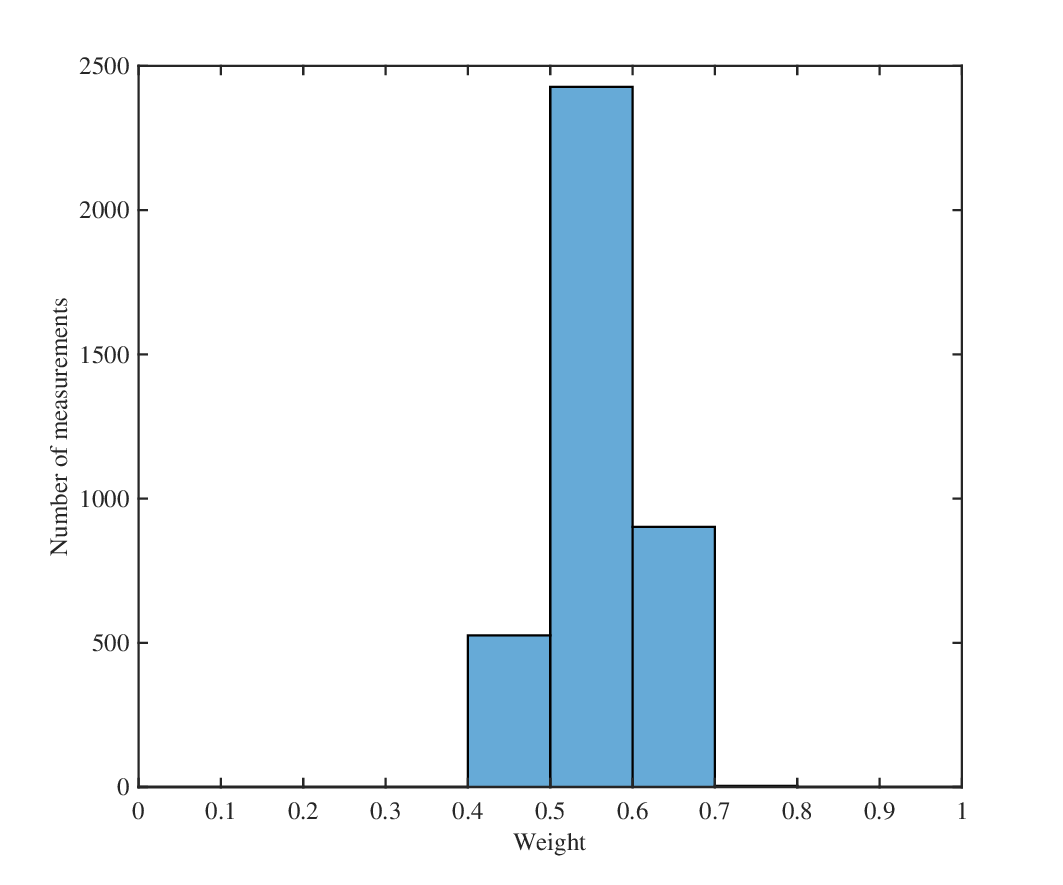}%
\label{subfig:linear}}
\hfil
\subfloat[]{\includegraphics[width=3.1in]{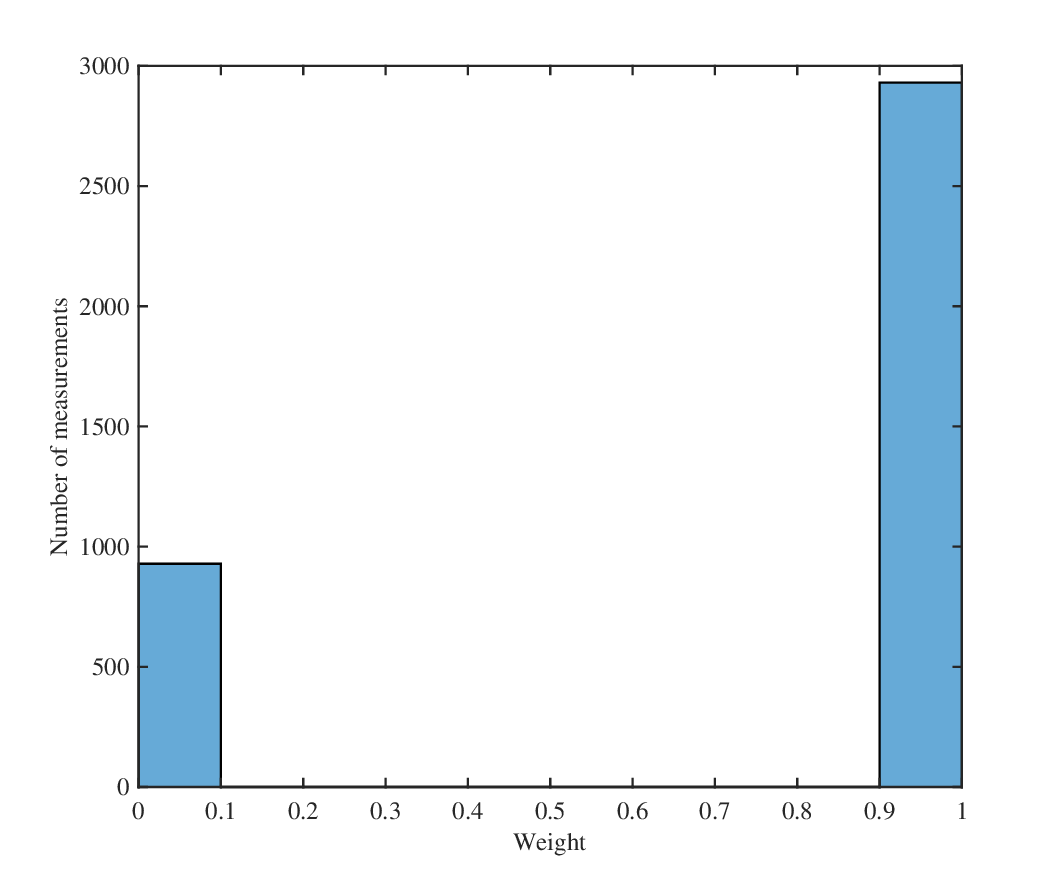}%
\label{subfig:step}}
\hfil
\subfloat[]{\includegraphics[width=3.1in]{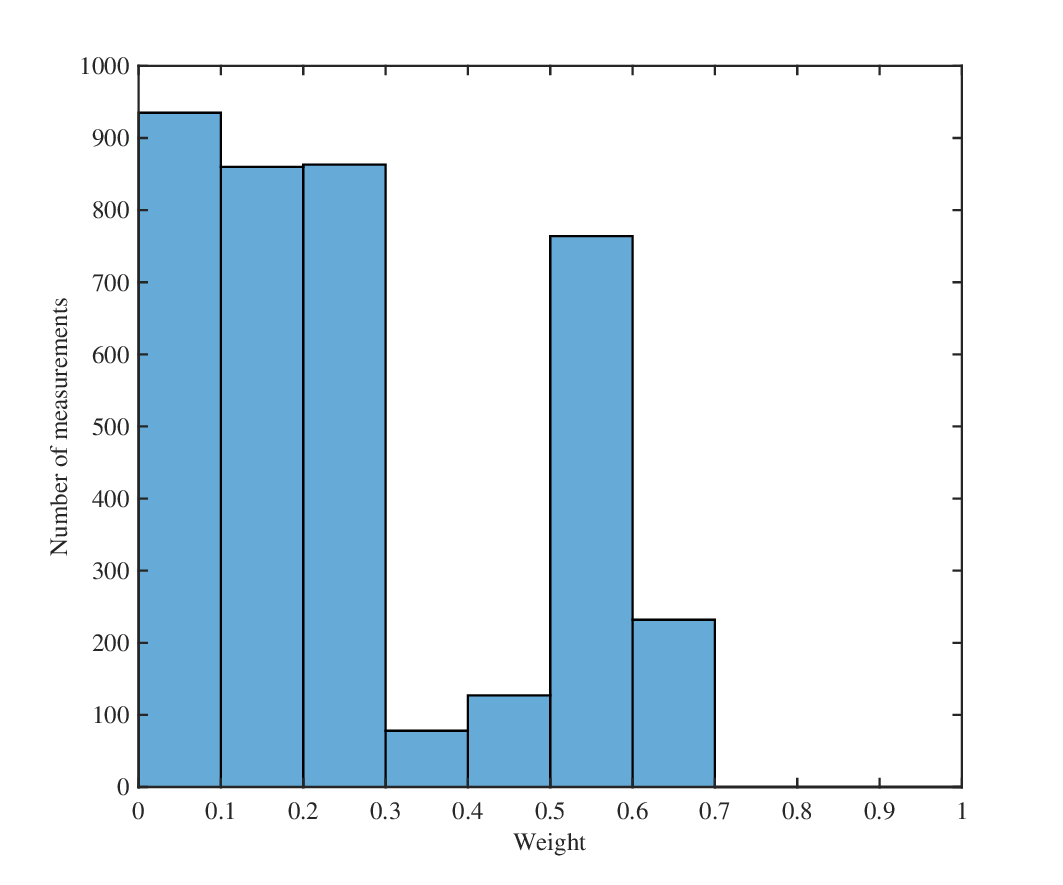}%
\label{subfig:relu}}
\hfil
\subfloat[]{\includegraphics[width=3.1in]{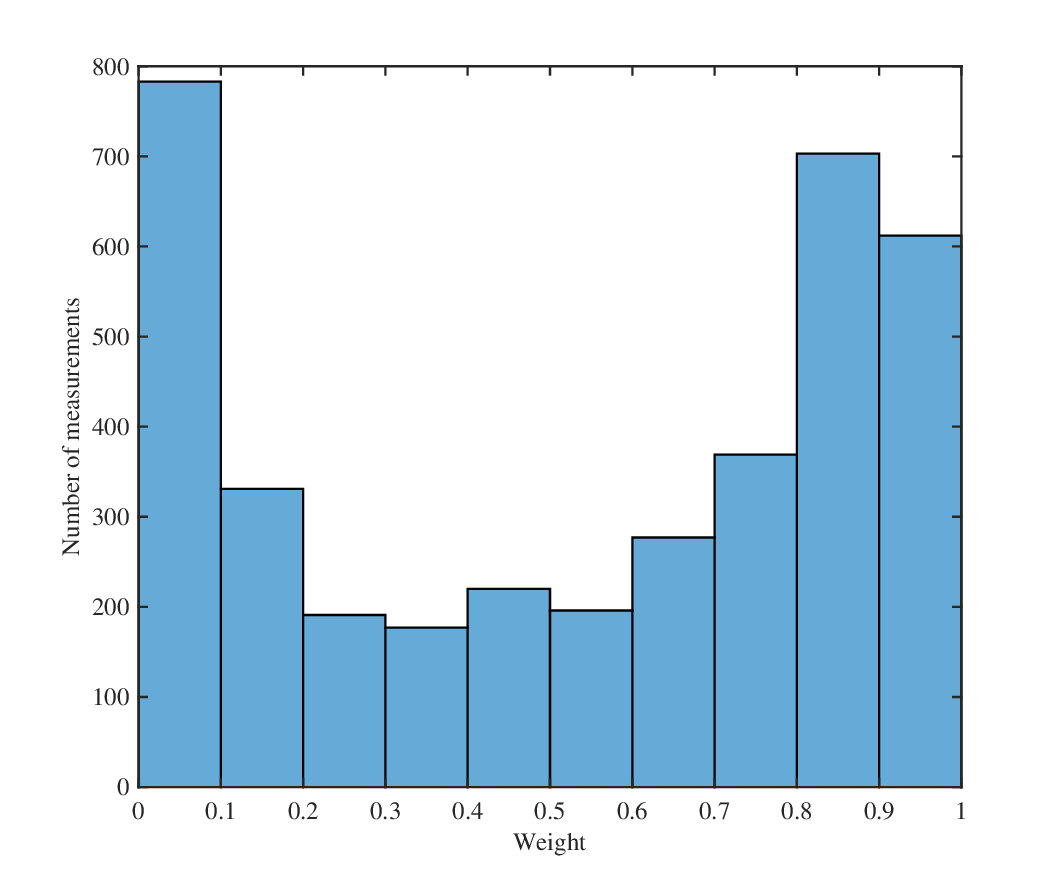}%
\label{subfig:sigmoid_50}}
\hfil
\subfloat[]{\includegraphics[width=3.1in]{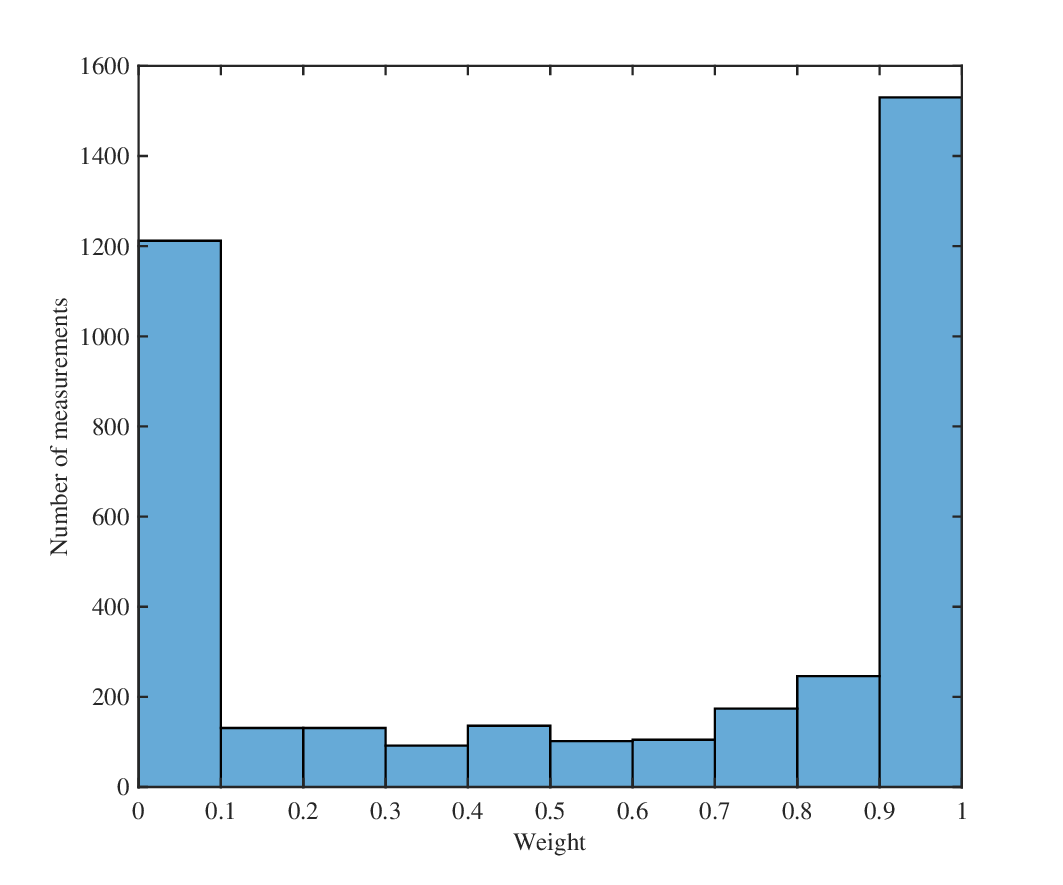}%
\label{subfig:sigmoid_100}}
\caption{Weight distribution of GPS data with activation functions: (a) constant (b) linear (c) unit step (d) ReLU (e) sigmoid a=mean b=50 (f) sigmoid a=mean b=100.}
\label{fig:activation_histogram}
\end{figure*}

In the proposed  framework, activation functions are used to map predicted scores by a machine learning algorithm to weights for WLS positioning calculation.. Figure \ref{fig:activation_histogram} presents the weight distribution of GPS testing data after applying the activation functions on AdaBoost predicted scores. Figure \ref{subfig:constant} presents the weight ditribution for the constant activation function of Figure  \ref{fig:activation_function} where the weights of all  signals are unity. Figure \ref{subfig:linear} employes the linear function from Figure \ref{fig:activation_function}  which directly assigns the predicted scores from AdaBoost results to WLS weights without any processing. The predicted scores are within a narrow interval between 0.4 to 0.7, and hence the use of this activation functionn results in a small variation among the assigned weights. When the weights become nearly identical, the WLS algorithm reduces to  OLS which diminishes the intended advantage of weighting. Figure \ref{subfig:step} illustrates the weight distribution corresponding to the unit step activation function of Figure  \ref{fig:activation_function}   where the threshold is set to the mean predicted scores for a given epoch. This case is equivalent to a satellite selection algorithms that calculates an OLS solution using satellite subsets. The weight distribution shows that nearly half of the measurements are excluded from producing solutions, that can seriously degrade the satellite geometry. The weight distribution of the ReLU function avoids deteriorating satellite geometry by eliminating only the signal with the lowest predicted score as shown in Figure \ref{subfig:relu}. The weights of remaining signals are linearly mapped to the range [0,1]. To further amplify the difference between weights of signals while excluding signals of extremely low quality, the sigmoid function is employed. In Figure \ref{subfig:sigmoid_50} and Figure \ref{subfig:sigmoid_100} we present the weight distribution of sigmoid functions with parameters $b=50$ and $b=100$, respectively. When the parameter $b$ increases, more measurements are pushed towards the extreme values  0 and 1 while the remaining signals have weights more evenly distributed between these two values. Rather than directly excluding low-quality signals, the sigmoid function gradually reduces their weights in accordance with their predicted quality. This soft approach preserves potentially useful measurements while still limiting their influence on the WLS solution. As a result, a sigmoid function can effectively reject poor quality signals while preserving satellite geometry. In addition, the widely spread weight distribution produced by a sigmoid function is preferred over a concentrated distribution produced by a linear function, since it enables WLS to better distinguish between high and low quality signals, thereby improving positioning accuracy. With all the advantages described, the sigmoid function can outperform other activation functions selected in this study. While ReLU might also be able to improve positioning accuracy by excluding the lowest quality signal, its improvement might not be as large compared to a sigmoid function due to a more concentrated weight distribution. In the following positioning results analysis, the constant function will serve as the baseline and results of ReLU and sigmoid functions will be compared.

\subsubsection{GPS-only Positioning Results}
The impact of the shaping parameter $b$ of the sigmoid function on positioning performance has been evaluated for values ranging from $b=1$ to $b=200$. Using the   random forest predicted scores, the optimal performance occurs at $b=27$ with a 3D RMSE of 167.627 (m). For the AdaBoost predicted scores, the minimum 3D RMSE of 164.803 (m) is achieved when $b=108$. The difference in the optimal values of parameter $b$ arises from the distinct predicted scores distributions produced by the random forest and AdaBoost algorithms. In the subsequent positioning analysis, the sigmoid function with the optimal values of parameter $b$ is used.

\begin{figure}[h!]
\centering
\includegraphics[width=6.2in]{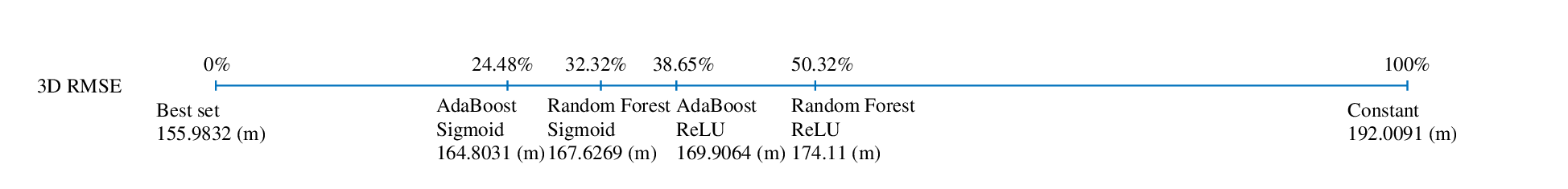}
\caption{3D RMSE of GPS-only positioning using Medium Urban data.}
\label{fig:GPS_only_RMSE}
\end{figure}

Figure \ref{fig:GPS_only_RMSE} presents the overall 3D RMSE of GPS-only positioning results. The baseline 3D RMSE produced by the constant activation function is around 192 (m). The best set refers to the satellite subset that yields the minimum 3D RMSE among all possible subset combinations. ``AdaBoost ReLU" and ``Random Forest ReLU" results are 61.35\% and 49.68\%  respectively from baseline towards the best set. ``AdaBoost Sigmoid" and ``Random Forest Sigmoid" results show an even larger improvement of 75.52\% and 67.68\%  respectively from the baseline towards the best set results. These results demonstrate that all the proposed methods effectively enhance positioning accuracy when using single  constellations, while AdaBoost performs better than random forest. The fact that the sigmoid function performs better than ReLU also aligns well with our activation function analysis. 

\begin{figure}[h!]
\centering
\includegraphics[width=4.1in]{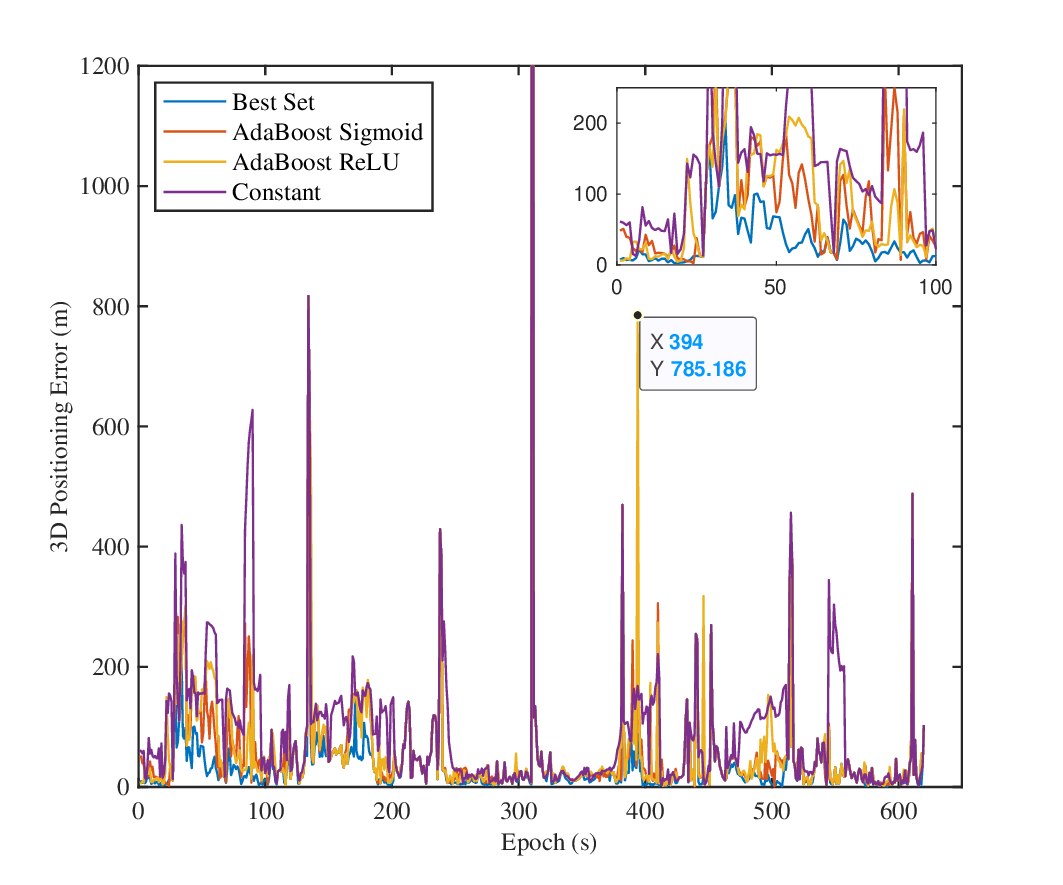}
\caption{3D errors of GPS-only positioning using Medium Urban data.}
\label{fig:GPS_3D_error}
\end{figure}

Figure \ref{fig:GPS_3D_error} presents the 3D positioning errors using AdaBoost predicted scores. This figure includes only the epochs with available positioning solutions, i.e., those with four or more signals available. The positioning errors produced by the ReLU and sigmoid activation functions are generally lower than those of the baseline. In certain epochs, the proposed methods achieve performance comparable to that of the best set. However, the ReLU curve shows more frequent error spikes and greater variability compared to the sigmoid. For example, ``AdaBoost ReLU" generates a peak error at epoch 394 which is caused by the inaccurate machine learning prediction. In this epoch, GPS space vehicle identifiers (SVID) 4, 7, 9, 21, and 27 are available and the best set includes GPS SVID 7, 9, 21, and 27. However, ReLU activation function assigns zero weight to good quality GPS SVID 27 signal and the remaining GPS SVID 4, 7, 9, 21 signals produce large positioning error. When only a few satellites are available, each one has a greater influence on the final WLS solution. In such cases, excluding a high-quality satellite signal and including a low-quality one can substantially reduce positioning accuracy.

\subsubsection{GPS+BeiDou Positioning Results}
\begin{figure}[h!]
\centering
\includegraphics[width=6.2in]{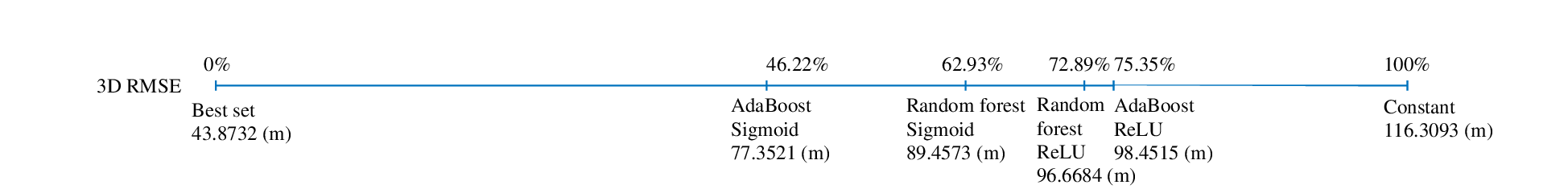}
\caption{3D RMSE of GPS+BeiDou positioning using Medium Urban data.}
\label{fig:Multi_RMSE}
\end{figure}

In multi-constellation positioning using GPS and BeiDou data, the optimal values of the  parameter $b$ in the sigmoid functions are $b=12$ and $b=120$ for random forest and AdaBoost predicted scores, respectively. Figure \ref{fig:Multi_RMSE} presents the 3D RMSE results for GPS+BeiDou positioning. It is seen that ``AdaBoost Sigmoid" achieves the best performance which improves the 3D RMSE by 53.78\% from baseline towards the optimal set. ``Random forest sigmoid" also demonstrates a moderate improvement of 37.07\% while ``Random forest ReLU" and ``AdaBoost ReLU" show similar improvement of around 26\% from baseline towards the best set. Using GPS and Beidou data together results in a lower improvement in 3D RMSE  compared to the that achieved in GPS-only positioning. This can be attributed to the relatively lower machine learning accuracy of BeiDou data compared to that of GPS data as shown in Table \ref{tbl:hk_gps_ml} and \ref{tbl:hk_bd_ml}. Multiple constellations do not aleays produce more accurate results than single constellations.

\begin{figure}[h!]
\centering
\includegraphics[width=4.1in]{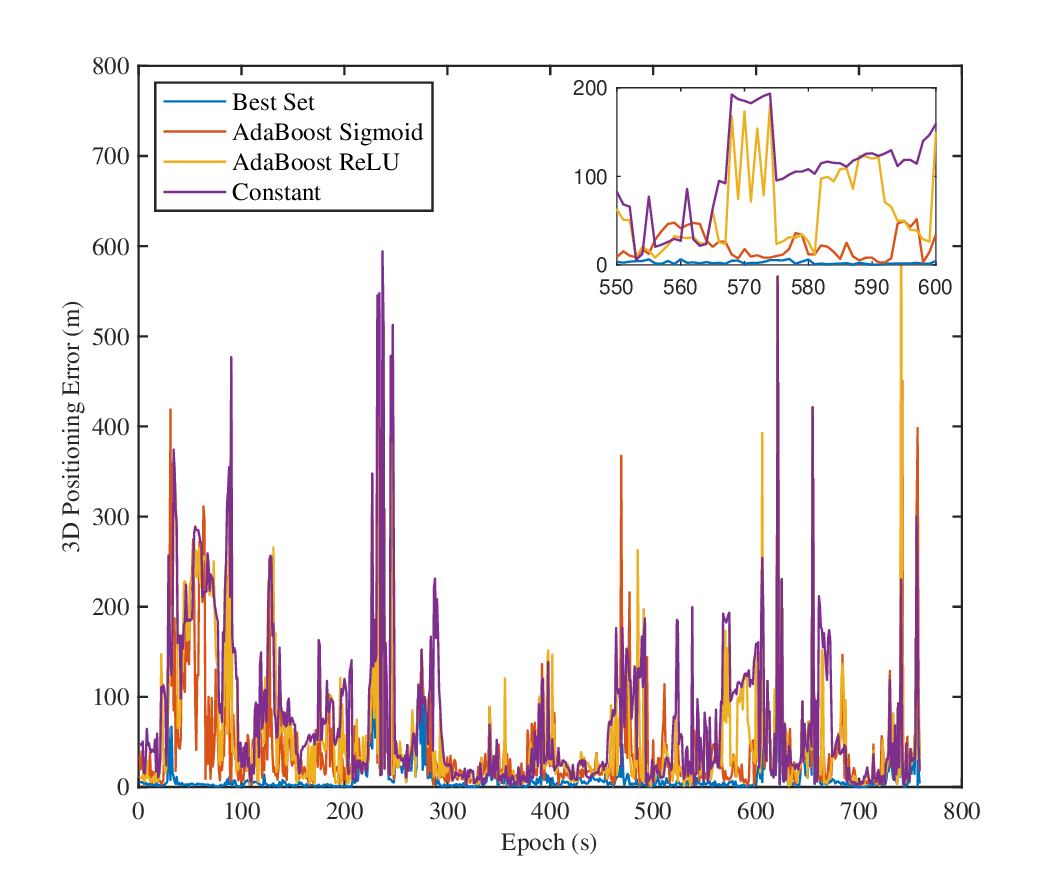}
\caption{3D errors of GPS+Beidou positioning using Medium Urban data.}
\label{fig:Multi_adaboost_3D_error}
\end{figure}

Figure \ref{fig:Multi_adaboost_3D_error} presents 3D errors of multi-constellation positioning results using AdaBoost predicted scores. The overall trends of ``AdaBoost Sigmoid" and ``AdaBoost ReLU" generate lower 3D positioning error than the baseline results. These results indicate that machine learning algorithms can effectively predict signals quality scores, while activation functions convert these predicted scores into suitable weights for WLS positioning that  reduce errors. Among the activation functions, ``AdaBoost Sigmoid" performs better than ``AdaBoost ReLU" across most of the epochs. Furthermore, as shown in the zoomed-in section in Figure \ref{fig:Multi_adaboost_3D_error}, the performance of ``AdaBoost Sigmoid" indicated by the orange line approaches that of best set indicated by the blue line. Since the ReLU activation function excludes only a single signal in each epoch, its impact on positioning performance may be limited when a large number of signals are available in multi-constellation positioning. In contrast, the sigmoid function can assign near-zero weights to multiple low-quality signals while amplifying the weight contrast between measurements to enhance the distinction between signals. These characteristics make sigmoid function more effective in enhancing  positioning accuracy. 

\subsubsection{Comparison of Single- and Multi-constellations}
\begin{figure}[h!]
\centering
\includegraphics[width=4.1in]{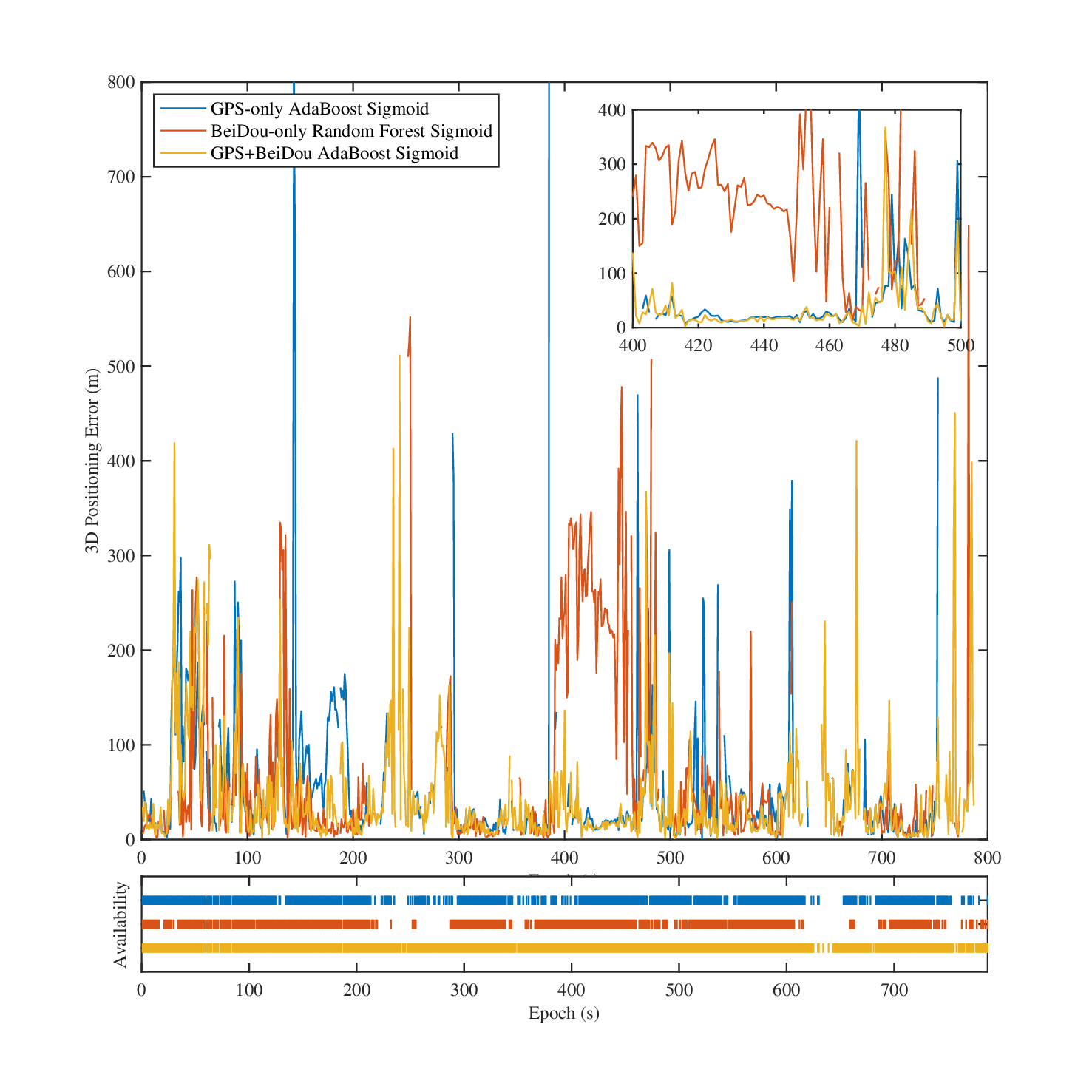}
\caption{Single and multi-constellation 3D positioning errors and time availabilities comparison.}
\label{fig:single_multi_compare}
\end{figure}

Figure \ref{fig:single_multi_compare} presents 3D positioning errors and time availability of GPS-only, BeiDou-only and GPS+BeiDou configurations with their respective best-performing combination of machine learning algorithm and activation function. The time availability results indicate that incorporating both GPS and BeiDou satellites enables position solutions to be obtained in nearly all epochs. This improvement stems from the increased number of available signals, which reduces the likelihood of encountering epochs with insufficient number of signal for a solution. In contrast, single-constellation GNSS configurations, such as GPS-only or BeiDou-only, experience longer and more frequent gaps in time availability. Despite multi-constellation GNSS providing positioning solutions when single-constellation systems lack sufficient signals for LS computation, the 3D errors of the positioning results are usually large, as observed in epochs 237–248. Such scenarios commonly involve significant signal obstruction and increased probability of NLOS/multipath effects. At epochs where positioning solutions are available for all  configurations, GPS+BeiDou results typically produce 3D errors that fall below or between the errors observed in GPS-only and BeiDou-only results. When one of these two constellation provide much worse results than the other constellation, using these two together will provide degraded results compared to using only the better constellation. Hence to exploit the two constellations in order to have results better than any single one, the performance of each one must be satisfactory.

\subsection{Results of Shinjuku Data Using Odaiba Training Set}
The Tokyo dataset is utilized to assess the generalization ability of the proposed algorithm across different geographical locations. In the following analysis, Odaiba data are used for machine learning training and Shinjuku data are used for testing. We focus exclusively on the best-performing AdaBoost algorithm for evaluation.

\subsubsection{Machine Learning Results}
Table \ref{tbl:odaiba_ml_adaboost} summarizes the training and testing accuracies of AdaBoost. The training accuracy is approximately 10\% higher than the testing accuracy for both GPS and BeiDou data. This can be caused by the environment differences between the training data and testing data. The Odaiba dataset includes mainly measurements collected in open areas, where NLOS and multipath errors are less likely to occur while the Shinjuku dataset contains signals collected in dense urban environments with a higher possibility of such errors. Consequently, in this case machine learning models do not generalize well to detect degraded signals, leading to a noticeable reduction in accuracy from training to testing.

\begin{table}[h!]
\caption{Machine learning results with Odaiba training set and Shinjuku testing set.}
\label{tbl:odaiba_ml_adaboost}
\centering
\begin{tabular}{|c|c|c|}
\hline
AdaBoost & GPS & BeiDou \\ \hline
Training Accuracy &  0.7910   &    0.7551 \\ \hline
Testing Accuracy  & 0.6999   &     0.6596 \\ \hline
\end{tabular}
\end{table}

\subsubsection{GPS+BeiDou Positioning Results}
\begin{figure}[h!]
\centering
\includegraphics[width=6.2in]{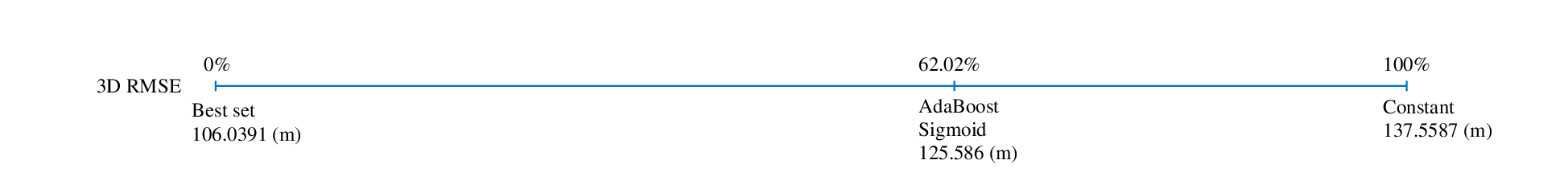}
\caption{3D RMSE of GPS+BeiDou positioning using Shinjuku data.}
\label{fig:Odaiba_Shinjuku_Multi_3D_RMSE}
\end{figure}

The best parameter of the sigmoid function is $b=47$ for GPS+BeiDou Shinjuku data and it results in a 3D RMSE of 125.586 (m). The 3D RMSE of the proposed algorithm with respect to baseline and best set is shown in Figure \ref{fig:Odaiba_Shinjuku_Multi_3D_RMSE}. The ``AdaBoost Sigmoid" algorithm produces a 3D RMSE which is 62.02\% away from the best set. While this represents an improvement, it is less significant compared to the previous cases, where the gap from the best set was reduced to 46.22\% for GPS+BeiDou positioning and 24.48\% for GPS-only positioning, relative to the baseline, in medium urban Hong Kong data. This can be the result of the lower machine learning testing accuracy. The AdaBoost testing accuracies of Shinjuku GPS data and BeiDou data in Table \ref{tbl:odaiba_ml_adaboost} are 6.7\% and 4.47\% lower than those in Hong Kong medium urban dataset in Table \ref{tbl:hk_gps_ml} and \ref{tbl:hk_bd_ml}. The 3D RMSE of ``AdaBoost ReLU" is not shown in Figure \ref{fig:Odaiba_Shinjuku_Multi_3D_RMSE} as it reaches 236.13 (m) which is 1.7 times larger than the baseline 3D RMSE. This is due to the presence of multiple large error peaks, one of which reaches 8440.47 (m).

\begin{figure}[h!]
\centering
\includegraphics[width=4.1in]{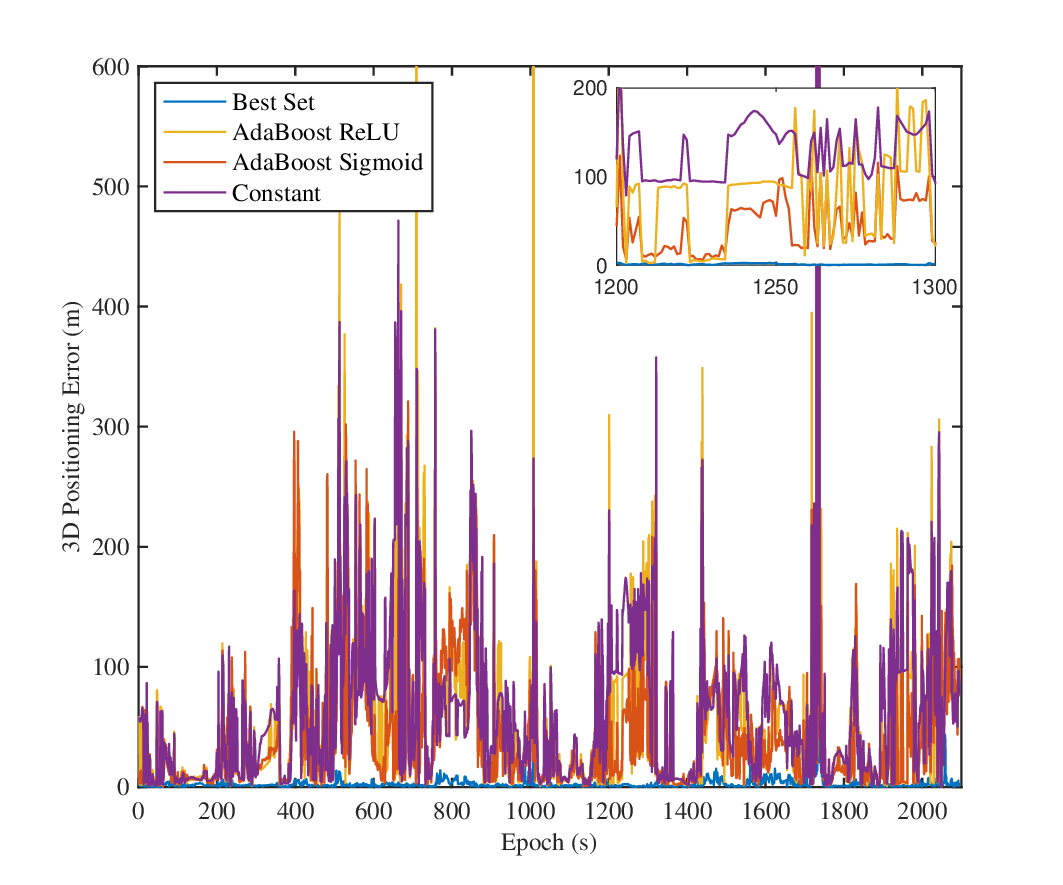}
\caption{3D errors of GPS-only positioning using Shinjuku data.}
\label{fig:Odaiba_Shinjuku_Multi_adaboost_3D_error}
\end{figure}

Figure \ref{fig:Odaiba_Shinjuku_Multi_adaboost_3D_error} presents 3D errors of multi-constellation positioning results using AdaBoost predicted scores. For most epochs, ``AdaBoost Sigmoid" and ``AdaBoost ReLU" yield smaller 3D positioning errors compared to the baseline. However, peak errors occur more frequently with the ``AdaBoost Sigmoid" and ``AdaBoost ReLU" methods, during which their positioning errors exceed the baseline errors. This outcome may be explained by the inaccurate machine learning predictions due to relatively low testing accuracy. For example, between epochs 771 and 843, the ``AdaBoost Sigmoid" and ``AdaBoost ReLU" methods produce higher 3D errors than the baseline. In epoch 801, the available signals are from GPS SVID 10, 12, 15, 20, 24, 25, 32 and BeiDou SVID 6, 7, 9, 16, 24, 25. The LS solution using all available satellites yields a 3D error of 73.6675 (m). The best set includes GPS SVID 10, 15 and BeiDou SVID 6, 7, 16 which generates a 3D error of 4.7112 (m). The ``AdaBoost Sigmoid" method assigns weights above 0.6 to GPS SVIDs 10, 20, and 25, as well as BeiDou SVIDs 7, 16, and 24. This weight distribution is inconsistent with the best set which leads to a 3D RMSE of nearly 139 (m). This epoch demonstrates the influence of machine learning prediction accuracy on the resulting positioning errors.


\subsection{Results of Shinjuku Data Using Hong Kong Training Set}
To assess the applicability of the proposed method when the machine learning model is trained in a different geographical location with similar urbanization levels, the Hong Kong training set and the Shinjuku testing set are employed in the following analysis. Our analysis focuses on the performance of the proposed method using the best-performing AdaBoost algorithm.

\subsubsection{Machine Learning Results}
The training and testing accuracies of AdaBoost algorithm are presented in Table \ref{tbl:hk_shinjuku_ml_adaboost}. The testing accuracy improves by 4.24\% for GPS data and 6.95\% for BeiDou data compared to the results in Table \ref{tbl:odaiba_ml_adaboost} where the Odaiba dataset is used as the training set. This improvement suggests that training a machine learning model on data collected in an urban environment similar to the testing conditions can effectively enhance the testing accuracy. Furthermore, the AdaBoost algorithm achieves similar training and testing accuracies which suggests that the model generalizes well to testing data. The accuracy  problem observed when using the Odaiba training set is not present when the model is trained on the Hong Kong dataset.

\begin{table}[h!]
\caption{Machine Learning Results with Hong Kong training set and Shinjuku testing set.}
\label{tbl:hk_shinjuku_ml_adaboost}
\centering
\begin{tabular}{|c|c|c|}
\hline
AdaBoost & GPS & BeiDou \\ \hline
Training Accuracy & 0.7603 &    0.7133 \\ \hline
Testing Accuracy & 0.7424 &     0.7291 \\ \hline
\end{tabular}
\end{table}

\subsubsection{GPS+BeiDou Positioning Results}
\begin{figure}[h!]
\centering
\includegraphics[width=6.2in]{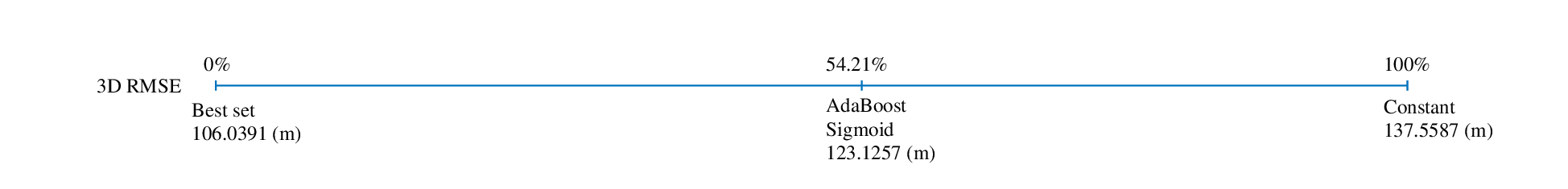}
\caption{3D RMSE of GPS+BeiDou positioning using Shinjuku data.}
\label{fig:HK_Shinjuku_Multi_3D_RMSE}
\end{figure}

The optimal value of for the sigmoid function parameter is $b=57$ which produces a 3D RMSE of 123.1257 (m). Figure \ref{fig:HK_Shinjuku_Multi_3D_RMSE} illustrates the 3D RMSE of proposed methods. The 3D RMSE of ``AdaBoost Sigmoid" method is 54.21\% away from the best set relative to the baseline. This represents an improvement of 7.81\% closer to the best set compared with the results obtained when using the Odaiba dataset for training. This result highlights the benefit of training the model with data collected in environments of comparable urbanization levels. Despite the geographical distance between Hong Kong and Tokyo, both Hong Kong training data and Shinjuku testing data are collected in dense urban environments, which leads to improved machine learning generalization and enhanced positioning performance. These results demonstrate that the proposed algorithms are applicable across different geographical locations. However, ``AdaBoost ReLU" method is not shown in the figure as it results in a 3D RMSE of 143.8039 (m) which exceeds the baseline 3D RMSE. This can be attributed to the peak errors that are larger than 3000 (m) in 4 of the epochs.

\begin{figure}[h!]
\centering
\includegraphics[width=4.1in]{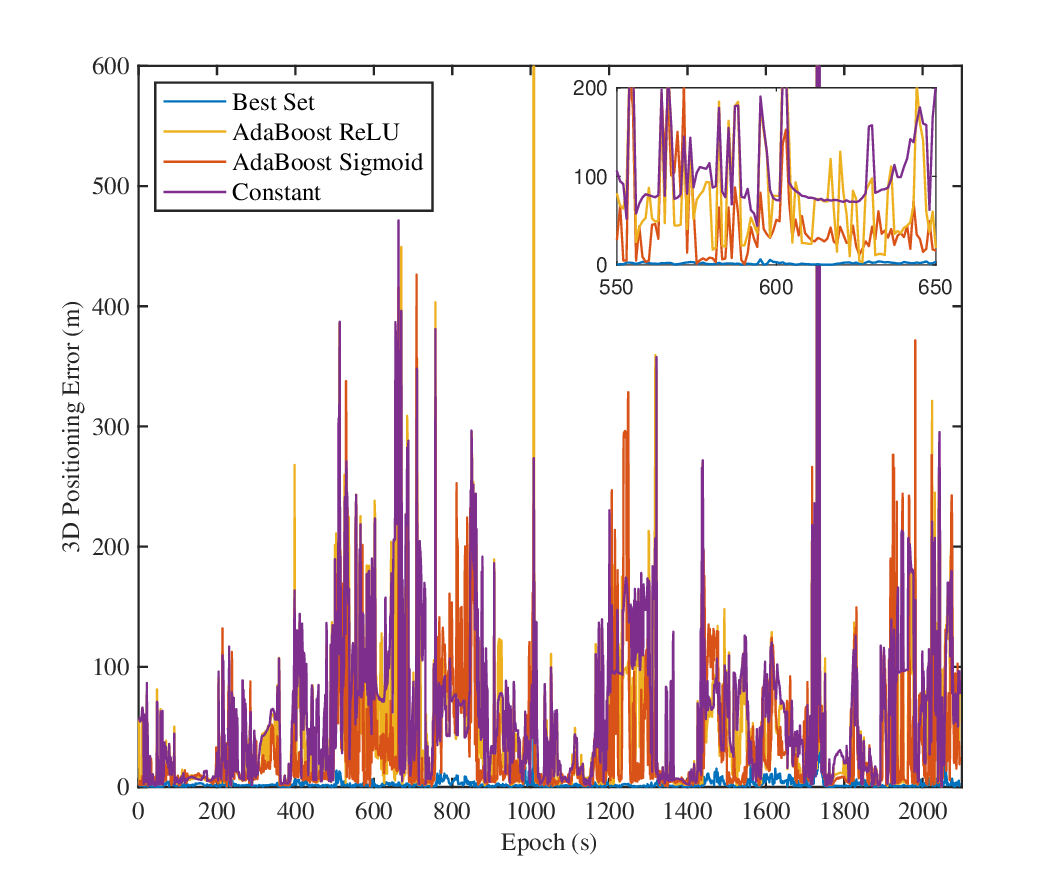}
\caption{3D errors of GPS-only positioning using Shinjuku data.}
\label{fig:HK_Shinjuku_Multi_3D_error}
\end{figure}

Figure \ref{fig:HK_Shinjuku_Multi_3D_error} presents 3D errors of GPS+BeiDou positioning using AdaBoost predicted scores. Overall, the ``AdaBoost Sigmoid" and ``AdaBoost ReLU" methods achieve lower 3D errors than the baseline which demonstrates both the effectiveness of the proposed algorithm and its applicability when the training data are obtained from geographically distant locations relative to the testing data. Compared to Figure \ref{fig:Odaiba_Shinjuku_Multi_adaboost_3D_error}, we can observe that the 3D errors of ``AdaBoost Sigmoid" in Figure \ref{fig:HK_Shinjuku_Multi_3D_error} are lower. The differences between these two results demonstrate the importance of selecting training data and its impact on positioning accuracy. However, there are still epochs where the proposed algorithms results in  larger 3D errors than the baseline due to the inaccurate machine learning prediction.

The results from Hong Kong training data show that the proposed method does not require training with datasets from every new site. Instead, a model trained on data collected in an environment with a similar level of urbanization is applicable to other locations under comparable conditions. This characteristic reduces the need of location-specific datasets, allowing the overall dataset size to remain relatively small. 


\section{Conclusion}\label{sec:conclusion}
This study proposes ensemble-based machine learning-aided WLS algorithms within a novel activation function framework. Six signal quality indicators are employed as machine learning input features to differentiate the quality of measurements. Ensemble machine learning algorithms, including random forest, AdaBoost and gradient boosting, are used to learn the relationship between selected features and optimal satellite subset. The activation functions are then used to transform the predicted scores from machine learning output into appropriate weights for WLS positioning. The proposed algorithms are evaluated using real-world datasets from two different locations, Hong Kong and Tokyo. Firstly, Hong Kong harsh urban and deep urban data are used for training  and Hong Kong medium urban data are used for testing. We demonstrate that AdaBoost algorithm performs better than random forest and sigmoid function outperforms ReLU. The 3D RMSE of best performing ``AdaBoost Sigmoid" results are 24.48\% and 46.32\% away from that of the best set for GPS-only and GPS+BeiDou positioning, respectively. This shows the proposed algorithms are effective in assigninig proper weigths to degraded signals in challenging urban environments which lead to improved positioning accuracy. Secondly, we conducted the experiment using Odaiba training set and Shinjuku testing set. The 3D RMSE of ``AdaBoost Sigmoid" result is 62.02\% away from best set for GPS+BeiDou positioning. Due to the environment difference in training and testing data, the improvement is not as significant as previous case. Lastly, the Hong Kong training data and Shinjuku testing data are used, both of which were collected in dense urban areas despite their different geographical locations. The 3D RMSE of ``AdaBoost Sigmoid" result is 54.21\% away from best set for GPS+BeiDou positioning. In comparison with the previous case, this result proves that training with data collected from environments that share a similar level of urbanization with the testing data can lead to better performance. 

Future work includes the integration of GNSS with inertial navigation systems (INS) using techniques such as Kalman filtering and factor graph optimization within the proposed machine learning–based activation function framework to further enhance positioning accuracy and time availability in challenging environments. The shape of the used activation function is critical in obtaining good results. In our work we show that Sigmoid activation functions provide good results for the data sets we ued. An interesting future research direction could be  the use of  machine learning techniques to adaptively select (or build) optimal activation functions, matching the parameters   to the environment and overall signal quality.

\section*{Acknowledgements}
The GNSS datasets that are  used in this paper are obtained from open-source UrbanNav repository collected by the
Intelligent Positioning and Navigation Laboratory. UrbanNav datasets can be downloaded
from https://github.com/IPNL-POLYU/UrbanNavDataset.
The authors thank the members of the Intelligent Positioning
and Navigation Laboratory led by Dr. Li-Ta Hsu from The
Hong Kong Polytechnique University for collecting and making
UrbanNav dataset available.

	\bibliographystyle{IEEEtran}
	\bibliography{references}

\end{document}